\documentclass[11pt]{article}

\usepackage[final]{acl}

\usepackage{times}
\usepackage{latexsym}

\usepackage[T1]{fontenc}

\usepackage[utf8]{inputenc}

\usepackage{microtype}
\usepackage{hyperref}
\usepackage{booktabs}
\usepackage{xcolor}
\usepackage{colortbl}
\usepackage{multirow}
\usepackage{amsmath}
\usepackage{siunitx}
\usepackage[most]{tcolorbox}
\tcbuselibrary{skins,breakable,minted}
\usepackage{listings}
\usepackage{makecell}
\usepackage{array}
\usepackage{enumitem}
\usepackage{amssymb}
\usepackage{graphicx}
\usepackage[noabbrev,capitalize]{cleveref}
\usepackage{fontawesome5}
\usepackage{seqsplit}

\usepackage{inconsolata}

\newcolumntype{L}[1]{>{\raggedright\arraybackslash}m{#1}}
\newcolumntype{C}[1]{>{\centering\arraybackslash}m{#1}}

\definecolor{goodred}{RGB}{220,53,69}
\definecolor{goodgreen}{RGB}{40,167,69}
\definecolor{rtfwd}{RGB}{37,99,235}
\definecolor{rtbwd}{RGB}{234,88,12}

\definecolor{py-bg}{HTML}{F8F9FB}
\definecolor{py-frame}{HTML}{DDE1E6}
\definecolor{py-kw}{HTML}{2563EB}      %
\definecolor{py-fn}{HTML}{7C3AED}      %
\definecolor{py-str}{HTML}{047857}     %
\definecolor{py-cmt}{HTML}{6B7280}     %
\definecolor{py-num}{HTML}{B91C1C}     %

\newcommand{\cmark}{\textcolor{green!70!black}{\ensuremath{\checkmark}}}
\newcommand{\xmark}{\textcolor{red!70!black}{\ensuremath{\times}}}

\lstdefinestyle{py-light}{
  language=Python,
  basicstyle=\ttfamily\small,
  backgroundcolor=\color{py-bg},
  frame=single, framerule=0.4pt, rulecolor=\color{py-frame},
  numbers=none, numberstyle=\tiny\color{py-cmt}, numbersep=8pt,
  keywordstyle=\color{py-kw},
  ndkeywordstyle=\color{py-fn},            %
  emphstyle=\color{py-fn},        %
  stringstyle=\color{py-str},
  commentstyle=\color{py-cmt},
  identifierstyle=\color{black},
  morekeywords={self,True,False,None},     %
  alsoletter={_},
  showstringspaces=false, upquote=true,
  breaklines=true, tabsize=2, columns=fullflexible,
  keepspaces=true
}

\newcommand{\invenc}{\mathsf{enc}^{-1}}
\newcommand{\invdec}{\mathsf{dec}^{-1}}
\newcommand{\enc}{\mathsf{enc}}
\newcommand{\dec}{\mathsf{dec}}

\newcommand{\datasetname}{\textsc{RoundTripCodeEval}} %

\newcommand{\avgcell}[1]{%
  \begingroup
  \def\val{#1}%
  \ifdim \val pt < 20pt
    \cellcolor{goodred!30}{#1}%
  \else\ifdim \val pt < 40pt
    \cellcolor{orange!30}{#1}%
  \else\ifdim \val pt < 60pt
    \cellcolor{yellow!35}{#1}%
  \else\ifdim \val pt < 80pt
    \cellcolor{goodgreen!25}{#1}%
  \else
    \cellcolor{goodgreen!40}{#1}%
  \fi\fi\fi\fi
  \endgroup
}

\NewDocumentCommand\emojirepeat{}{\,{\tiny\textcolor{rtfwd}{\faSync}}}

\title{Can LLMs Compress (and Decompress)?\\ Evaluating Code Understanding and Execution via Invertibility}

\author{Nickil Maveli \quad Antonio Vergari\textsuperscript{\emojirepeat} \quad Shay B. Cohen\textsuperscript{\emojirepeat} \\
  School of Informatics, University of Edinburgh \\
  10 Crichton Street, Edinburgh, EH8 9AB \\
  \texttt{\{nickil.maveli,avergari,scohen\}@ed.ac.uk}}

\begin{document}
\maketitle

\begin{abstract}
LLMs demonstrate strong performance on code benchmarks, yet consistent reasoning across forward and backward execution remains elusive. We present~\datasetname{} (RTCE), a benchmark of four code execution reasoning tasks that evaluates round-trip consistency through execution-free, exact-match assessment of bijection fidelity across four lossless compression algorithms. We evaluate state-of-the-art Code-LLMs under zero-shot prompting, supervised fine-tuning on execution traces, and iterative self-reflection. All approaches yield only modest improvements and none closes the gap, revealing that current LLMs lack the internal coherence required for reliable bidirectional code reasoning. RTCE surfaces findings invisible to existing benchmarks: models frequently pass individual forward and backward tasks yet fail the combined round-trip, exposing mutually inconsistent internal representations; SFT and self-reflection saturate after one revision round, indicating they cannot repair fundamental algorithmic misunderstandings; and failures persist even on simple bijections such as RLE, suggesting that algorithmic complexity is not the sole root cause.\footnote{Code and dataset are available at \url{https://github.com/Nickil21/round-trip-code-compression}.
{\emojirepeat} denotes shared supervision.}
\end{abstract}

\section{Introduction}
Can LLMs understand code and algorithms? What would it mean for them to understand code and algorithms to begin with? We frame such understanding as a \emph{code inversion} problem. Indeed, recent progress in Code-LLMs~\citep{zheng-etal-2024-opencodeinterpreter, lozhkov2024starcoder2stackv2, hui2024qwen25codertechnicalreport, guo2024deepseekcoderlargelanguagemodel} has demonstrated remarkable performance across various software engineering applications. However, evaluating the reasoning ability of Code-LLMs requires going beyond isolated input–output predictions. Most existing code reasoning benchmarks evaluate single-direction execution, either forward execution, i.e., predicting the output from the input and code~\citep{pmlr-v235-gu24c,jain2025livecodebench,11029885,li2025codeio,liu2024codemind} or backward execution, i.e., inferring the input from the output and code~\citep{pmlr-v235-gu24c,li2025codeio}. While these tasks are valuable, they overlook a key property of robust reasoning systems: the ability to integrate forward and backward execution into a coherent and reversible process~\citep{jiang-etal-2024-forward}. Forward execution can often be solved through surface-level pattern matching, memorisation, or statistical correlation rather than genuine mechanistic understanding~\citep{pmlr-v235-gu24c,jain2025livecodebench}. Inversion, however, is fundamentally different. It requires the model to understand the forward execution as a bijective encoding function and then construct a corresponding decoding function that perfectly recovers the original input. An LLM might achieve high accuracy in one direction, yet fail to maintain logical compatibility when the process is inverted, leading to inconsistencies~\citep{min2024beyond,pmlr-v235-allamanis24a,xu-etal-2025-cruxeval,liu2025assessing}. Success on a round-trip inversion provides more substantial evidence of deep semantic code understanding than forward-only accuracy. Because inversion cannot be solved by local pattern matching alone, models must implicitly construct a consistent internal execution model. Failures to close the loop and achieve self-consistency thus indicate that forward correctness was fragile, derived from template matching and API memorisation~\citep{wang-etal-2024-unlocking} rather than mechanistic reasoning~\citep{he-etal-2025-large} about data flow and control logic. Investigation into these areas is vital to advance LLMs from pattern-based generation tools to mechanistically grounded, trustworthy code assistants capable of deep, bidirectional understanding.

\begin{figure}
    \centering
    \includegraphics[width=1.\linewidth]{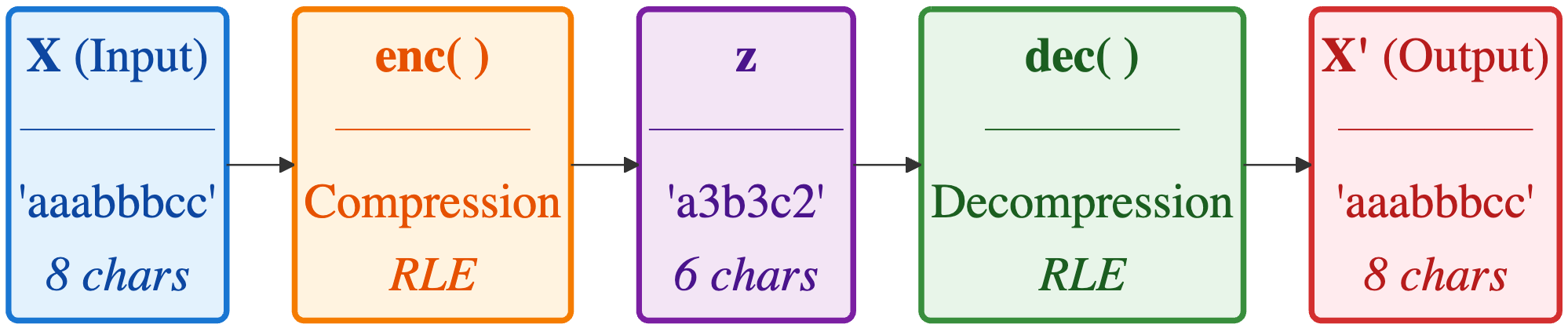}
    \caption{A standard lossless compression pipeline, where an input \(X\) is encoded into a compact representation \(z\) and then decoded back into \(X'\). Here, \texttt{enc} and \texttt{dec} can be referred to as the encoding and decoding functions. We outline the overall workflow of our pipeline using a concrete example in Appendix~\ref{sec:task-example}.
    }
    \label{fig:basic-compression-pipe}
\end{figure}

To address this gap, we introduce a round-trip code evaluation framework where the model must apply a transformation (encoding) and its inverse (decoding) such that the final reconstruction is identical to the original input. The transformation is a bijection because it is both injective, mapping each input to a unique encoded output, and surjective, covering the entire encoded space so that decoding perfectly recovers the original input. The inherent invertibility and enforced self-consistency offer a robust and comprehensive diagnostic for LLM reasoning capabilities. By simultaneously testing the completeness of the forward mapping and the accuracy of the inverse mapping, this framework exposes subtle asymmetries, inconsistencies, and reasoning failures that are undetectable by conventional single-direction evaluations. Through this closed-loop evaluation, we move beyond isolated input–output correctness, instead quantifying a model's internal coherence, bidirectional reasoning strength, and reliability, which are indispensable in code-based applications that demand high precision, transparency, and robustness.

We construct RTCE in a controlled setting with 250 inputs per algorithm drawn from diverse real-world sources, providing a rigorous measure of LLM reasoning in challenging scenarios. The benchmark's core challenge is the inversion task, which goes beyond the forward execution traces on which LLMs are primarily trained. The forward task is essentially a code-completion or execution task familiar to the model; the inverse task forces the model to treat the given function's output as input and infer the original semantics to reverse the operation. We evaluate models under zero-shot prompting to establish a baseline, supervised fine-tuning on execution traces to maximise the forward signal, and iterative self-reflection to test whether models can self-correct systematic errors. Together, these three paradigms determine whether any current approach can close the round-trip gap, or whether the failure is fundamental to how models internalise code semantics.

\section{Problem Statement}
Code invertibility, in the context of programming, refers to the ability to reverse a piece of code to its original state or to retrieve the initial input from the output. In this work, we study code invertibility by framing it as a \emph{round-trip} through a lossless, symmetric compression–decompression pipeline as illustrated in Figure~\ref{fig:basic-compression-pipe}. We evaluate it through the lens of self-consistency, the requirement that ensures re-encoding any decoded input reproduces the original representation exactly.

Formally, let $x \in \mathcal{X}$ be the original input string, 
$z \in \mathcal{Z}$ its encoded compressed output,
and $x' \in \mathcal{X}$ the reconstruction after decoding.
Define the encoding and decoding maps
\[
\enc \colon \mathcal{X} \to \mathcal{Z},
\qquad
\dec \colon \mathcal{Z} \to \mathcal{X}.
\]
We write
\[
z = \enc(x), \qquad x' = \dec(z),
\]
subject to the \emph{lossless round-trip} constraint
\[
\forall\, x \in \mathcal{X}, \quad \dec(\enc(x)) = x.
\]

\begin{figure*}[t]
    \centering
    \includegraphics[width=1.\linewidth]{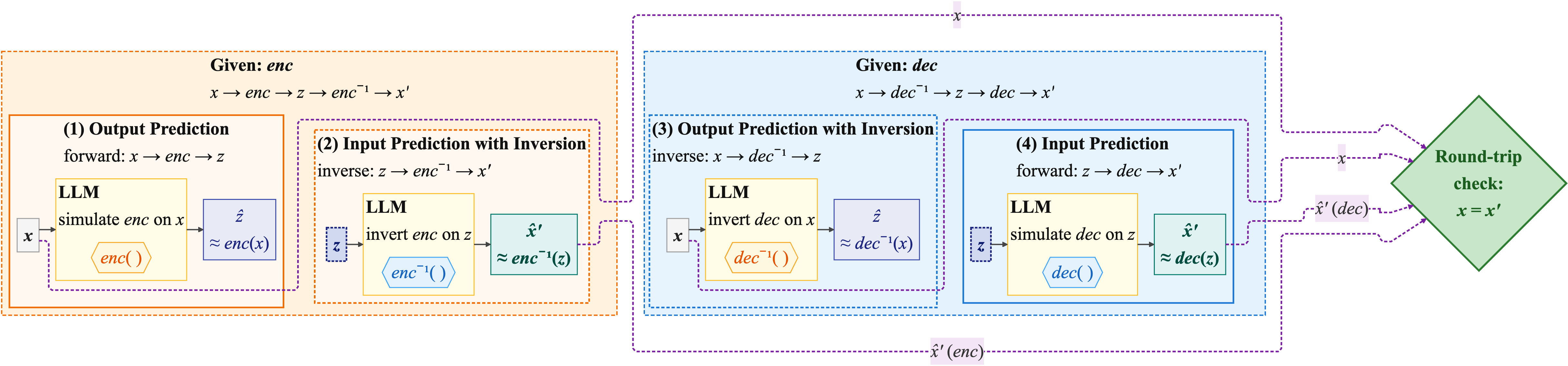}
    \caption{
Overview of the reasoning tasks which depict our four-step round-trip procedure for assessing code compression self-consistency. In (1), \textbf{Output Prediction}, the input is transformed using the actual encoder \( {\scriptstyle x \xrightarrow{\enc} z} \) and the LLM predicts \(\hat{z}\) from \(x\) and \(\enc\). In (2), \textbf{Input Prediction \emph{with Inversion}}, the compressed output is mapped back to the input space \( {\scriptstyle z \xrightarrow{\invenc} x'} \) using the inverted encoder \(\invenc\) to reconstruct \(\hat{x}'\). In (3), \textbf{Output Prediction \emph{with Inversion}}, the input is passed through the inverted decoder \( {\scriptstyle x \xrightarrow{\invdec} z} \) to recover \(\hat{z}\). In (4), \textbf{Input Prediction}, the compressed output is decoded \( {\scriptstyle z \xrightarrow{\dec} x'} \) using the actual decoder \(\dec\) to produce the final reconstruction \(\hat{x}'\). A round-trip check \( {\scriptstyle x {=} x'} \) verifies perfect reconstruction fidelity, returning a binary result that indicates if the LLM is self-consistent or not. This corresponds to a chain of length 1, but this could be extended to an arbitrary length.
    }
    \label{fig:task-overview}
\end{figure*}

Unlike round-trip tasks involving natural language translation (e.g., code $\to$ description $\to$ code), which only require semantic equivalence (the final code behaves the same), lossless compression requires exact data identity. This is a much stricter, unambiguous test of mechanistic reasoning. Within this framework, we delineate four closely related prediction tasks as illustrated in Figure~\ref{fig:task-overview}:

\begin{itemize}[noitemsep,parsep=0pt,leftmargin=*]
    \item \textbf{Output Prediction:} Given $(x,\ \enc)$, predict $\hat{z} \approx \enc(x)$.
    This measures forward reasoning and checks the LLM's ability to simulate the compression process. We denote it by \( {\scriptstyle x \xrightarrow{\enc} z} \).

    \item \textbf{Input Prediction \emph{with Inversion}:} Given $(z,\ \enc)$, predict $\hat{x}' \approx \enc^{-1}(z)\ \equiv\ \dec(z)$.
    This evaluates backward reasoning and checks whether the LLM can internally reinterpret the decoder as its inverse encoder to reconstruct the input. We denote it by \( {\scriptstyle z \xrightarrow{\invenc} x'} \).

    \item \textbf{Output Prediction \emph{with Inversion}:} Given $(x,\ \dec)$, predict $\hat{z} \approx \dec^{-1}(x)\ \equiv\ \enc(x)$.
      This evaluates backward reasoning and checks whether the LLM can internally reinterpret the encoder as its inverse decoder to obtain the compressed output. We denote it by \( {\scriptstyle x \xrightarrow{\invdec} z} \).

    \item \textbf{Input Prediction:} Given $(z,\ \dec)$, predict $\hat{x}' \approx \dec(z)$.
    This tests forward reasoning and checks the LLM's ability to simulate the decompression process. We denote it by \( {\scriptstyle z \xrightarrow{\dec} x'} \).
\end{itemize}

While there is a class of algorithms for which code inversion is possible (see \S\ref{section:benchmark}), code inversion is a non-trivial problem in the general case. Consider that any arbitrary algorithmic problem $A(x) = z$ could be changed into a function $B(x) = (x,z) = (x,A(x))$ (including the input in the output). Inverting $B(x)$ into $C(x,z)$ is algorithmically trivial (define $C$ as returning $x$ given $(x,z)$), but the inversion of $C$ is the arbitrary $A$. Therefore, for undecidable problems where a witness helps verify an input-output pair (such as a variant of the Post Correspondence problem, where $x$ is two equal-length lists of strings and $z$ is non-empty sublists of each list, with identical indices from the original lists, such that the two concatenations of strings in each sublist are the same), it is easy to construct $C(x,z) = x$ (while implementing a simple check to see whether $z$ is indeed a witness), but inverting $C$ is non-trivial (with the code for the simple check provided), because the original $A$, solving the PC problem (or providing a witness to a solution), does not exist as a normally executable algorithm.

\begin{table*}[ht]
\centering
\resizebox{0.7\linewidth}{!}{%
\begin{tabular}{lcccccl}
\toprule
\textbf{Benchmark} & \textbf{Fwd} & \textbf{Bwd} & \textbf{RT} & \textbf{Invert?} & \textbf{Exec?} & \textbf{Comparison Level} \\
\midrule
IdentityChain & \cmark & \cmark & Partial & \xmark & No  & Spec $\leftrightarrow$ Code \\
RTC           & \cmark & \cmark & Strict  & \xmark & No  & Code $\leftrightarrow$ NL descriptions \\
CodeIO        & \cmark & \cmark & No      & \xmark & Yes & Function I/O \\
CRUXEval      & \cmark & \cmark & No      & \xmark & Yes & Function exec I/O \\
CodeMind      & \cmark & \xmark & No      & \xmark & Yes & Task-level exec reasoning \\
ExeRScope     & \cmark & \xmark & No      & \xmark & Yes & Feature-level exec reasoning \\
REVAL         & \cmark & \xmark & No      & \xmark & Yes & Runtime behavior \\
CACP          & \cmark & \xmark & No      & \xmark & Yes & Concept-level traces \\
RTCE (ours)   & \cmark & \cmark & Strict  & \cmark & Yes & Exact enc/dec I/O bijection \\
\bottomrule
\end{tabular}%
}
\caption{Comparison between RTCE and previously released benchmarks. Here, RT: Round-trip.}
\label{tab:benchmark_comparison}
\end{table*}

\section{Related Work}
Existing benchmarks evaluate forward or backward execution in isolation; none ask whether both directions are mutually consistent. RTCE closes this loop, drawing on the following themes.
\paragraph{Self-consistency.}
The principle of self-consistency~\citep{wang2023selfconsistency} has recently gained traction in LLM evaluation, especially for code generation. The IdentityChain framework~\citep{min2024beyond} reveals that models often generate code and corresponding specifications that appear correct in isolation but become inconsistent when composed, exposing reasoning failures that conventional one-directional benchmarks overlook. Round-Trip Correctness (RTC;~\citealt{pmlr-v235-allamanis24a}) evaluates a model's ability to describe code in natural language and then regenerate the original code from that description. By enforcing consistency between the original and reconstructed code, RTC provides an unsupervised and scalable evaluation method that applies across real-world code domains without the need for costly human annotations. Unlike RTC, which tests semantic equivalence between code and its natural language description, and IdentityChain, which checks consistency across specification-code pairs, our closed-loop evaluation operates directly within program execution and enforces exact bijection rather than semantic equivalence.

\paragraph{Round-trip code evaluation.}
CodeIO~\citep{li2025codeio}, CRUXEval~\citep{pmlr-v235-gu24c} and its multilingual extension CRUXEval-X~\citep{xu-etal-2025-cruxeval} probe bidirectional input–output reasoning over executable Python functions and score each direction independently. CodeMind~\citep{liu2024codemind} and ExeRScope~\citep{liu2025toolindepthanalysiscode} check reasoning ability about program execution and require models to track control flow, state changes, and dependencies across program steps. REVAL~\citep{chen2024reasoningruntimebehaviorprogram} and CACP~\citep{10.5555/3692070.3692823} analyse intermediate reasoning traces during execution to expose concept-level misunderstandings and inconsistencies in LLMs. Our framework unifies and extends these directions under the notion of round-trip self-consistency, where a prediction is considered valid only if its logical inverse reconstructs the original input exactly. We do not rely on annotated traces or concept-level labelling; instead, by requiring predictions to be reconstructible through their logical inverse, our evaluation surfaces asymmetric reasoning failures that remain hidden under one-directional execution checks, trace-based diagnostics, or loosely paired evaluations.

\paragraph{Iterative self-correction.}
Several works have explored whether LLMs can improve their own outputs through iterative critique and revision~\citep{madaan2023selfrefine,asai2024selfrag,shinn2023reflexion,gou2024critic}. \citet{olausson2024repair} show that self-repair for code generation offers only marginal gains without grounding in external execution feedback, a finding that directly aligns with the saturation we observe after the first revision round in RTCE. Unlike these works, which target code synthesis, we apply iterative self-reflection as a diagnostic to probe whether models can recover round-trip consistency, demonstrating that the failure mode in RTCE is not addressable through self-correction alone.

Table~\ref{tab:benchmark_comparison} shows the comparison table that highlights what RTCE uniquely measures compared to previous works.

\section{Benchmark Construction}
\label{section:benchmark}

\begin{table*}[t]
\centering
\resizebox{0.8\linewidth}{!}{%
\begin{tabular}{p{4cm}cp{12cm}}
\toprule
\textbf{Data Source} & \textbf{\# Categories} & \textbf{Key Characteristics} \\
\midrule
Patterned String Data
& $15$ 
& Repeated characters, alternating/block patterns, palindromes and near-palindromes, pangrams, keyboard sequences, pseudo-random token insertions, and short natural language sentences (single and repeated). Mix of highly compressible patterns and high-entropy noise. \\
\midrule
Structured Log Data
& 11 
& Deterministic Apache-style log lines with timestamp, log level, HTTP method, endpoint, module, synthetic user ID, request duration, and IP. Categories include slow requests, status/metrics checks, authentication events, database errors, and severity levels. Metadata provided for analysis. \\
\midrule
YAML-Like Config Data
& 10 
& Compact but realistic configuration snippets: application configs, Kubernetes deployments, Docker Compose, Helm values, Ansible playbooks, Prometheus configs, GitHub Actions, CircleCI, CloudFormation, Terraform. Preserves indentation, key–value syntax, and multi-document formats. \\
\midrule
Tabular Data 
& 10 
& CSV/TSV tables (5$\times$5) with numeric, alphanumeric, mixed type, sparse, and repeated-header variants. Variations in delimiter, value entropy, and sparsity capture real-world spreadsheet/log regularity and redundancy. \\
\bottomrule
\end{tabular}%
}
\caption{Summary of the four synthetic data source families used in RTCE, including category counts and defining characteristics.}
\label{tab:data_sources}
\end{table*}

Unlike prevailing code-generation benchmarks that grade programs against hidden unit tests or online judges~\citep{chen2021codex,austin2021program}, we target exact algorithmic behaviour with a fully deterministic, execution-free evaluation. RTCE is built through a three-stage process: \emph{generation}, where synthetic inputs are created across four source families (patterned strings, structured logs, YAML-like configs, tabular data) with multiple subcategories to ensure diversity and balance representative of real-world coding and data processing tasks; \emph{validation}, where deterministic reference implementations of four compression algorithms, namely, Lempel–Ziv–Welch (LZW; \citealt{1659158}), Arithmetic Encoding (AE; \citealt{5391119}), Run-Length Encoding (RLE; \citealt{Golomb1966}) and Huffman~\citep{4051119}, produce exact ground-truth outputs under a fixed seed; and \emph{serialisation}, where validated input–output pairs are stored both as raw files by algorithm and as a unified JSON corpus containing descriptions, labels, and metadata. All random operations are governed by a fixed seed for reproducibility.

We select algorithms that span the four canonical design paradigms in lossless compression: dictionary, statistical interval coding, run aggregation, and prefix coding, while remaining simple enough to yield unique and deterministic ground truths. For LZW, the output is a sequence of integer codes representing substrings found during scanning. For AE, the output consists of a single floating-point value in the range $[0,1)$, denoting the midpoint of the final coding interval and a frequency dictionary containing counts for all symbols, including a special EOF marker, necessary for exact decoding. For RLE, the output is a list of $(\text{symbol}, \text{count})$ pairs. For Huffman Coding, the output includes a list of integers representing byte-packed Huffman-encoded bitstrings and a metadata JSON file containing the symbol-to-bitstring codebook and the padding length needed for decoding. Collectively, these algorithms span a spectrum of input complexities and compression strategies, making them highly suitable for evaluating the capacity of code LLMs to learn and invert diverse algorithmic transformations. Compression tasks involve iterative symbol substitutions, dictionary management, probabilistic coding, and bitwise encoding, demanding multi-step algorithmic reasoning and memory. Such algorithmic complexity exceeds the relatively arbitrary construction of many generic bijective functions.

RTCE includes 250 unique input samples, each assessed across four distinct code execution tasks, resulting in a comprehensive dataset of 1000 evaluation examples. Input lengths range from a few characters to several hundred, offering a diverse evaluation spectrum. We are in a controlled setting, and we do this on purpose to clearly measure whether LLMs can reliably perform code understanding and execution. RTCE uses synthetic data that deliberately mirrors common artefacts in real developer workflows, rather than arbitrary toy strings. Because compression algorithms are content-agnostic, they operate solely on sequences of characters/symbols and never inspect the input's syntax or semantics. For this reason, what matters is the distribution of character patterns, not whether the string came from a GitHub repository or a synthetic generator. Detailed category-level statistics are summarised in Table~\ref{tab:data_sources}. The precise implementations of the encoding and decoding functions for each compression algorithm are provided in Appendix~\ref{sec:compression-algo}.

\section{Experimental Setup}

We provide information about the models we use and our evaluation method.

\subsection{Models}
LLMs can be grouped by how they are trained and optimised, with each type having its strengths and weaknesses:
\paragraph{General instruction:} Broadly trained LLMs using diverse web-scale corpora spanning natural language, factual knowledge, and limited code, but lack the inductive biases required for highly structured, algorithm-sensitive reasoning.
\vspace{-0.5em}
\paragraph{Code generation:} LLMs fine-tuned or pre-trained on large-scale, high-quality programming datasets, enabling execution reliability across a wide range of code-related tasks.
\vspace{-0.5em}
\paragraph{Reasoning / Reasoning-distilled:} Architectures explicitly optimised for multi-step analytical reasoning, frequently distilled from larger reasoning-focused systems capable of structured problem decomposition while maintaining inference efficiency.

\subsection{Evaluation}
We report three complementary metrics across all tasks and models.

\textbf{Exact Match (EM)} counts a prediction as correct only when it is value-equivalent to the reference: integers exactly, floats via \texttt{isclose} ($\text{tol}=10^{-3}$), and lists/dicts recursively; strings are case-folded and trimmed before comparison. \textbf{Edit Similarity (ES)} is a normalised Levenshtein score that captures partial credit for structurally plausible but symbol-imprecise outputs. \textbf{Pass@5} generates $n{=}5$ completions per instance and scores an instance correct if at least one completion achieves EM, measuring a model's ceiling under sampling independently of single-shot reliability.

In our experiments, we answer three inter-dependent research questions, increasing the level of complexity of using the LLMs to solve RTCE:

\begin{tcolorbox}[
  enhanced, breakable,
  colback=blue!4!white, colframe=blue!40!black,
  boxrule=0.6pt, arc=3pt,
  left=6pt, right=6pt, top=4pt, bottom=4pt,
  title={\textbf{RQ1}}, fonttitle=\bfseries\small,
  coltitle=blue!40!black, attach boxed title to top left={yshift=-2pt, xshift=6pt},
  boxed title style={colback=blue!4!white, colframe=blue!40!black, boxrule=0.6pt, arc=2pt}
]
How well can LLMs invert code in a zero-shot setting, relying solely on their parametric knowledge without any task-specific prompting or training?
\end{tcolorbox}

\begin{tcolorbox}[
  enhanced, breakable,
  colback=blue!4!white, colframe=blue!40!black,
  boxrule=0.6pt, arc=3pt,
  left=6pt, right=6pt, top=4pt, bottom=4pt,
  title={\textbf{RQ2}}, fonttitle=\bfseries\small,
  coltitle=blue!40!black, attach boxed title to top left={yshift=-2pt, xshift=6pt},
  boxed title style={colback=blue!4!white, colframe=blue!40!black, boxrule=0.6pt, arc=2pt}
]
Can iterative self-reflection, where the model critiques and revises its own outputs, recover LLM performance on code inversion beyond what zero-shot prompting achieves?
\end{tcolorbox}

Through RQ1 and RQ2, we probe whether code invertibility emerges naturally in LLMs or can be surfaced through guided self-reflection. Our conclusion across both RQs is that code invertibility is a deeply challenging problem for current LLMs, one that neither scale nor self-reflection alone appears sufficient to resolve. This naturally raises the question of whether the capability can instead be instilled through targeted supervision and learning from examples.

\begin{tcolorbox}[
  enhanced, breakable,
  colback=blue!4!white, colframe=blue!40!black,
  boxrule=0.6pt, arc=3pt,
  left=6pt, right=6pt, top=4pt, bottom=4pt,
  title={\textbf{RQ3}}, fonttitle=\bfseries\small,
  coltitle=blue!40!black, attach boxed title to top left={yshift=-2pt, xshift=6pt},
  boxed title style={colback=blue!4!white, colframe=blue!40!black, boxrule=0.6pt, arc=2pt}
]
Can fine-tuning on execution traces give LLMs the ability to invert code, or do persistent failures suggest a reasoning limitation that standard training alone cannot overcome?
\end{tcolorbox}

\section{Results}

We turn to explore the results for the three posed RQs.

\subsection{RQ1: Zero-shot model performance}
We evaluate 15 LLMs spanning four size tiers (1B--33B) across four task types per algorithm under zero-shot prompting. The tasks cover both directions of the encoder--decoder duality: \textbf{O/P Pred} ($x \xrightarrow{\enc} z$) applies the encoder directly; \textbf{O/P Pred-I} ($x \xrightarrow{\invdec} z$) provides the decoder and asks the model to invert it; \textbf{I/P Pred} ($z \xrightarrow{\dec} x'$) applies the decoder directly; \textbf{I/P Pred-I} ($z \xrightarrow{\invenc} x'$) provides the encoder and asks the model to invert it. All variants are strictly zero-shot, receiving one worked example in the preamble but no task-specific demonstrations. Prompt templates, input examples, and the inference pipeline are in Appendix~\ref{sec:prompt-templates}--\ref{sec:model-inference}. Table~\ref{tab:model-performance} reports Pass@5 results across all 15 LLMs and four algorithms.

\begin{table*}[htbp]
\centering
\sisetup{detect-weight=true,mode=text,round-mode=places,round-precision=2}
\resizebox{1\linewidth}{!}{%
\begin{tabular}{l c l S S S S S S S S S S S S r}
\toprule
\multicolumn{1}{c}{\textbf{Model}} &
\multicolumn{1}{c}{\textbf{Size (B)}} &
\multicolumn{1}{c}{\textbf{Alg.}} &
\multicolumn{3}{c}{\textbf{O/P Pred ($\mathbf{x} \xrightarrow{\enc} \mathbf{z}$)}} &
\multicolumn{3}{c}{\textbf{O/P Pred-I ($\mathbf{x} \xrightarrow{\invdec} \mathbf{z}$)}} &
\multicolumn{3}{c}{\textbf{I/P Pred ($\mathbf{z} \xrightarrow{\dec} \mathbf{x}'$)}} &
\multicolumn{3}{c}{\textbf{I/P Pred-I ($\mathbf{z} \xrightarrow{\invenc} \mathbf{x}'$)}} &
\multicolumn{1}{c}{\textbf{Average}} \\
\cmidrule(lr){4-6}
\cmidrule(lr){7-9}
\cmidrule(lr){10-12}
\cmidrule(lr){13-15}
& & & \textbf{EM} & \textbf{ES} & \textbf{P@5}
& \textbf{EM} & \textbf{ES} & \textbf{P@5}
& \textbf{EM} & \textbf{ES} & \textbf{P@5}
& \textbf{EM} & \textbf{ES} & \textbf{P@5}
& \\
\midrule
\multirow{4}{*}{Llama-3.2-1B-Instruct} & \multirow{4}{*}{1.0}
& AE & 0.00 & 0.69 & 0.00 & 0.00 & 1.62 & 0.00 & 0.00 & 0.16 & 0.00 & 0.00 & 1.58 & 0.00 & \avgcell{0.34} \\
& & LZW & 0.00 & 0.03 & 0.00 & 0.00 & 0.08 & 0.00 & 0.00 & 0.11 & 0.00 & 0.00 & 0.35 & 0.00 & \avgcell{0.05} \\
& & RLE & 0.00 & 0.20 & 0.00 & 0.00 & 0.05 & 0.00 & 0.00 & 0.81 & 0.00 & 0.00 & 0.72 & 0.00 & \avgcell{0.15} \\
& & HUFF & 0.00 & 0.00 & 0.00 & 0.00 & 0.00 & 0.00 & 0.00 & 0.27 & 0.00 & 0.00 & 0.68 & 0.00 & \avgcell{0.08} \\
\midrule
\multirow{4}{*}{DeepSeek\_R1\_Distill\_Qwen\_1.5B} & \multirow{4}{*}{1.5}
& AE & 0.72 & 15.00 & 2.00 & 0.00 & 1.00 & 0.00 & 0.00 & 2.71 & 0.00 & 0.00 & 0.53 & 0.00 & \avgcell{1.83} \\
& & LZW & 0.00 & 1.12 & 0.00 & 0.00 & 0.33 & 0.00 & 0.00 & 2.01 & 0.00 & 0.00 & 0.38 & 0.00 & \avgcell{0.32} \\
& & RLE & 2.80 & 12.32 & 7.20 & 0.40 & 0.81 & 2.00 & 1.60 & 7.44 & 3.60 & 1.84 & 6.78 & 4.80 & \avgcell{4.30} \\
& & HUFF & 0.00 & 0.62 & 0.00 & 0.00 & 0.01 & 0.00 & 0.16 & 5.78 & 0.40 & 0.00 & 0.65 & 0.00 & \avgcell{0.64} \\
\midrule
\multirow{4}{*}{Phi\_3\_mini\_128k\_instruct} & \multirow{4}{*}{3.8}
& AE & 0.00 & 6.85 & 0.00 & 0.00 & 0.91 & 0.00 & 0.08 & 15.38 & 0.40 & 0.00 & 7.55 & 0.00 & \avgcell{2.60} \\
& & LZW & 0.00 & 9.48 & 0.00 & 0.08 & 8.66 & 0.40 & 0.24 & 12.83 & 1.20 & 0.00 & 10.87 & 0.00 & \avgcell{3.65} \\
& & RLE & 1.84 & 41.86 & 2.80 & 0.40 & 18.21 & 2.00 & 3.76 & 31.40 & 7.20 & 2.64 & 26.43 & 5.60 & \avgcell{12.01} \\
& & HUFF & 0.00 & 0.66 & 0.00 & 0.00 & 0.93 & 0.00 & 0.00 & 10.04 & 0.00 & 0.00 & 6.88 & 0.00 & \avgcell{1.54} \\
\midrule
\multirow{4}{*}{Phi\_3.5\_mini\_instruct} & \multirow{4}{*}{3.8}
& AE & 0.56 & 24.00 & 1.20 & 0.00 & 1.07 & 0.00 & 0.00 & 5.55 & 0.00 & 0.00 & 1.78 & 0.00 & \avgcell{2.85} \\
& & LZW & 0.00 & 16.41 & 0.00 & 0.00 & 7.54 & 0.00 & 0.00 & 15.94 & 0.00 & 0.08 & 17.38 & 0.40 & \avgcell{4.81} \\
& & RLE & 1.84 & 53.66 & 4.00 & 0.00 & 0.41 & 0.00 & 4.00 & 45.72 & 7.20 & 3.52 & 37.27 & 6.40 & \avgcell{13.67} \\
& & HUFF & 0.00 & 0.51 & 0.00 & 0.08 & 0.26 & 0.40 & 0.08 & 11.29 & 0.40 & 0.00 & 7.37 & 0.00 & \avgcell{1.70} \\
\midrule
\multirow{4}{*}{Mistral\_7B\_Instruct\_v0.3} & \multirow{4}{*}{7.2}
& AE & 0.00 & 7.92 & 0.00 & 0.00 & 7.99 & 0.00 & 0.00 & 7.79 & 0.00 & 0.00 & 0.96 & 0.00 & \avgcell{2.05} \\
& & LZW & 0.00 & 9.85 & 0.00 & 0.00 & 5.68 & 0.00 & 0.00 & 9.29 & 0.00 & 0.00 & 2.14 & 0.00 & \avgcell{2.25} \\
& & RLE & 0.64 & 44.06 & 2.00 & 1.60 & 32.36 & 3.60 & 0.56 & 35.20 & 1.20 & 0.16 & 16.02 & 0.80 & \avgcell{11.52} \\
& & HUFF & 0.00 & 0.62 & 0.00 & 0.00 & 0.39 & 0.00 & 0.00 & 9.22 & 0.00 & 0.00 & 1.62 & 0.00 & \avgcell{0.99} \\
\midrule
\multirow{4}{*}{Qwen2.5\_7B\_Instruct} & \multirow{4}{*}{7.6}
& AE & 0.72 & 30.68 & 1.20 & 0.56 & 28.27 & 0.80 & 0.00 & 10.56 & 0.00 & 0.08 & 5.34 & 0.40 & \avgcell{6.55} \\
& & LZW & 0.00 & 6.31 & 0.00 & 0.16 & 9.98 & 0.80 & 0.16 & 16.36 & 0.40 & 0.40 & 18.20 & 0.80 & \avgcell{4.46} \\
& & RLE & 2.96 & 34.74 & 5.20 & 2.88 & 31.51 & 5.60 & 4.96 & 51.88 & 6.00 & 3.76 & 52.81 & 6.40 & \avgcell{17.39} \\
& & HUFF & 0.00 & 0.07 & 0.00 & 0.00 & 0.46 & 0.00 & 0.00 & 15.59 & 0.00 & 0.16 & 14.77 & 0.80 & \avgcell{2.65} \\
\midrule
\multirow{4}{*}{Llama\_3.1\_8B\_Instruct} & \multirow{4}{*}{8.0}
& AE & 0.72 & 15.71 & 0.80 & 0.40 & 19.02 & 0.80 & 0.32 & 7.84 & 0.80 & 0.08 & 6.47 & 0.40 & \avgcell{4.45} \\
& & LZW & 0.00 & 3.42 & 0.00 & 0.00 & 1.60 & 0.00 & 0.00 & 3.64 & 0.00 & 0.00 & 1.21 & 0.00 & \avgcell{0.82} \\
& & RLE & 2.08 & 26.02 & 4.40 & 1.60 & 12.52 & 3.60 & 2.72 & 29.95 & 5.20 & 0.24 & 23.23 & 0.80 & \avgcell{9.36} \\
& & HUFF & 0.00 & 0.00 & 0.00 & 0.00 & 0.16 & 0.00 & 0.00 & 10.22 & 0.00 & 0.00 & 3.18 & 0.00 & \avgcell{1.13} \\
\midrule
\multirow{4}{*}{codegemma\_7b\_it} & \multirow{4}{*}{8.5}
& AE & 0.16 & 13.67 & 0.80 & 0.08 & 6.37 & 0.40 & 0.00 & 5.81 & 0.00 & 0.08 & 2.92 & 0.40 & \avgcell{2.56} \\
& & LZW & 0.16 & 41.71 & 0.80 & 0.16 & 11.40 & 0.80 & 0.00 & 9.12 & 0.00 & 0.00 & 3.98 & 0.00 & \avgcell{5.68} \\
& & RLE & 1.68 & 54.53 & 3.20 & 0.24 & 5.93 & 1.20 & 0.40 & 14.54 & 1.20 & 0.48 & 3.22 & 1.60 & \avgcell{7.35} \\
& & HUFF & 0.00 & 0.98 & 0.00 & 0.00 & 1.11 & 0.00 & 0.00 & 8.97 & 0.00 & 0.00 & 2.50 & 0.00 & \avgcell{1.13} \\
\midrule
\multirow{4}{*}{Yi\_Coder\_9B\_Chat} & \multirow{4}{*}{8.8}
& AE & 0.08 & 6.24 & 0.40 & 0.00 & 0.84 & 0.00 & 0.00 & 11.86 & 0.00 & 0.00 & 7.04 & 0.00 & \avgcell{2.20} \\
& & LZW & 1.20 & 14.35 & 2.40 & 0.24 & 1.64 & 1.20 & 0.40 & 15.51 & 1.20 & 0.32 & 10.87 & 1.20 & \avgcell{4.21} \\
& & RLE & 10.56 & 37.44 & 14.00 & 2.40 & 5.02 & 9.20 & 5.76 & 32.89 & 8.00 & 0.80 & 13.33 & 2.80 & \avgcell{11.85} \\
& & HUFF & 0.00 & 0.13 & 0.00 & 0.00 & 0.00 & 0.00 & 0.00 & 10.64 & 0.00 & 0.00 & 11.63 & 0.00 & \avgcell{1.87} \\
\midrule
\multirow{4}{*}{DeepSeek\_R1\_Distill\_Qwen\_14B} & \multirow{4}{*}{14.8}
& AE & 4.88 & 32.12 & 11.20 & 5.20 & 31.05 & 13.20 & 0.40 & 9.99 & 1.20 & 0.64 & 9.92 & 1.20 & \avgcell{10.08} \\
& & LZW & 8.24 & 28.78 & 14.00 & 7.52 & 32.60 & 13.60 & 2.24 & 22.38 & 5.60 & 2.56 & 24.40 & 6.40 & \avgcell{14.03} \\
& & RLE & 16.16 & 63.25 & 26.00 & 16.88 & 56.03 & 28.00 & 8.08 & 47.25 & 13.20 & 4.56 & 27.39 & 16.80 & \avgcell{26.97} \\
& & HUFF & 0.00 & 0.06 & 0.00 & 0.00 & 0.45 & 0.00 & 0.68 & 15.59 & 1.51 & 1.13 & 15.69 & 2.63 & \avgcell{3.15} \\
\midrule
\multirow{4}{*}{Codestral\_22B\_v0.1} & \multirow{4}{*}{22.2}
& AE & 0.00 & 2.20 & 0.00 & 0.00 & 0.80 & 0.00 & 0.00 & 11.33 & 0.00 & 0.00 & 6.78 & 0.00 & \avgcell{1.76} \\
& & LZW & 0.88 & 23.35 & 1.20 & 0.64 & 23.94 & 1.20 & 1.36 & 22.52 & 2.40 & 0.32 & 14.23 & 1.20 & \avgcell{7.77} \\
& & RLE & 9.92 & 69.62 & 13.60 & 8.32 & 70.80 & 13.20 & 11.60 & 71.34 & 20.40 & 11.04 & 47.49 & 20.80 & \avgcell{30.68} \\
& & HUFF & 0.00 & 0.08 & 0.00 & 0.00 & 0.10 & 0.00 & 0.00 & 6.26 & 0.00 & 0.00 & 11.57 & 0.00 & \avgcell{1.50} \\
\midrule
\multirow{4}{*}{QwQ\_32B} & \multirow{4}{*}{32.8}
& AE & 12.64 & 23.06 & 27.59 & 18.27 & 33.81 & 41.62 & 1.85 & 11.01 & 2.31 & 2.20 & 11.32 & 2.89 & \avgcell{15.71} \\
& & LZW & 15.52 & 36.67 & 18.03 & 16.15 & 44.62 & 18.68 & 9.18 & 45.08 & 12.57 & 9.34 & 48.46 & 15.38 & \avgcell{24.14} \\
& & RLE & 32.00 & 80.23 & 57.60 & 36.80 & 78.62 & 66.80 & 26.08 & 79.49 & 53.20 & 34.96 & 79.76 & 61.20 & \avgcell{57.23} \\
& & HUFF & 0.00 & 0.00 & 0.00 & 0.00 & 0.22 & 0.00 & 4.34 & 18.25 & 7.89 & 4.71 & 19.52 & 11.11 & \avgcell{5.50} \\
\midrule
\multirow{4}{*}{Qwen2.5\_Coder\_32B\_Instruct} & \multirow{4}{*}{32.8}
& AE & 0.24 & 34.03 & 0.80 & 0.24 & 37.51 & 1.20 & 0.24 & 15.65 & 1.20 & 0.08 & 9.77 & 0.40 & \avgcell{8.45} \\
& & LZW & 4.56 & 63.76 & 7.60 & 2.88 & 66.75 & 7.20 & 2.16 & 46.34 & 3.60 & 2.08 & 41.79 & 4.00 & \avgcell{21.06} \\
& & RLE & 20.00 & 83.93 & 26.40 & 18.64 & 79.84 & 28.40 & 23.84 & 77.72 & 41.60 & 10.56 & 53.97 & 33.20 & \avgcell{41.51} \\
& & HUFF & 0.00 & 0.00 & 0.00 & 0.00 & 0.41 & 0.00 & 1.38 & 19.31 & 3.85 & 0.46 & 11.57 & 0.76 & \avgcell{3.15} \\
\midrule
\multirow{4}{*}{DeepSeek\_R1\_Distill\_Qwen\_32B} & \multirow{4}{*}{32.8}
& AE & 9.36 & 29.77 & 21.05 & 12.71 & 21.55 & 30.59 & 1.40 & 11.20 & 2.34 & 1.18 & 8.16 & 3.53 & \avgcell{12.74} \\
& & LZW & 12.50 & 64.39 & 17.61 & 13.14 & 50.77 & 20.57 & 5.23 & 35.09 & 10.23 & 6.29 & 36.23 & 13.71 & \avgcell{23.81} \\
& & RLE & 23.39 & 74.52 & 39.13 & 19.56 & 42.14 & 44.10 & 15.28 & 63.93 & 33.62 & 11.79 & 29.62 & 39.30 & \avgcell{36.37} \\
& & HUFF & 0.00 & 0.00 & 0.00 & 0.00 & 0.09 & 0.00 & 2.81 & 16.80 & 7.41 & 2.84 & 11.88 & 5.97 & \avgcell{3.98} \\
\midrule
\multirow{4}{*}{deepseek\_coder\_33b\_instruct} & \multirow{4}{*}{33.3}
& AE & 0.48 & 12.00 & 0.80 & 0.00 & 1.55 & 0.00 & 0.24 & 14.87 & 0.80 & 0.00 & 9.29 & 0.00 & \avgcell{3.34} \\
& & LZW & 0.08 & 2.86 & 0.40 & 0.08 & 3.52 & 0.40 & 0.48 & 18.85 & 1.20 & 0.24 & 12.40 & 0.80 & \avgcell{3.44} \\
& & RLE & 7.20 & 52.72 & 9.60 & 0.24 & 3.79 & 0.80 & 4.80 & 54.46 & 6.80 & 1.12 & 18.62 & 4.40 & \avgcell{13.71} \\
& & HUFF & 0.00 & 0.10 & 0.00 & 0.00 & 0.00 & 0.00 & 0.00 & 10.88 & 0.00 & 0.00 & 3.56 & 0.00 & \avgcell{1.21} \\
\bottomrule
\end{tabular}
}%
\caption{Unified comparison across models present in all four algorithm tables (AE, LZW, RLE, HUFFMAN). The colour-coded ``Average'' column (red = low, green = high) aggregates all the EM, ES, and Pass@5 (P@5) metrics, enabling quick visual comparison of model performance, strengths, and trends. Across all settings, results show a consistent hierarchy: {O/P Pred $>$ O/P Pred-I $>$ I/P Pred $>$ I/P Pred-I}. O/P Pred scores highest, and it seems that forward encoding an input string from compression is easier for most LLMs. I/P Pred-I and O/P Pred-I are the most challenging, as they combine strict input and output demands with inverse decoding, adding complexity and compounding errors.}
\label{tab:model-performance}
\end{table*}

\paragraph{Scale and algorithm difficulty.}
Models $\leq$3.8B score near zero across all algorithms, lacking the capacity for multi-step bookkeeping. The 7--9B range generalises on RLE (locally repetitive patterns) but fails on AE and LZW (long-span interval accumulation and dictionary management). Above 14B, RLE and LZW improve substantially, yet Huffman encoding remains unsolved for all 15 models. Reasoning-distilled models (e.g., DeepSeek-R1-14B; \citealt{Guo_2025}) consistently outperform general-instruction counterparts of comparable size, confirming that chain-of-thought pretraining helps with structured multi-step simulation.

\paragraph{Encoding--decoding asymmetry.}
Encoding (O/P Pred) requires tracking the evolving state across every symbol (interval bounds for AE, a growing dictionary for LZW, run-length counters for RLE), so even a single bookkeeping error can collapse exact matches. Decoding (I/P Pred) allows the model to exploit surface regularities in the encoded string, and I/P Pred Pass@5 exceeds O/P Pred for most pairs. AE is the exception: QwQ-32B reaches 27.6\% on AE encoding but only 2.3\% on AE decoding (a 12$\times$ collapse), because decoding requires inverse floating-point interval arithmetic that is more numerically fragile than the forward pass.

\paragraph{Inverse variants and the Huffman paradox.}
On RLE and AE, QwQ-32B~\citep{qwen2.5} scores higher on O/P Pred-I (66.8 and 41.6 Pass@5) than O/P Pred (57.6 and 27.6), because the provided decoder function is simpler to invert than the encoder is to simulate directly. This advantage vanishes for LZW and Huffman. More strikingly, all 15 models score 0\% on Huffman encoding (O/P Pred, O/P Pred-I) while QwQ-32B reaches 7.9\% on Huffman decoding (I/P Pred) and 11.1\% on I/P Pred-I. Decoding is a tree traversal over an explicitly provided tree; encoding requires constructing the frequency table, building the Huffman tree, and emitting variable-length codes: a multi-stage hierarchical procedure no model handles correctly. Finally, Edit Similarity remains above zero even at 0\% exact match (e.g., Mistral-7B on AE: ES $\approx 8\%$, P@5 $= 0\%$), showing that models produce plausible but imprecise outputs, and that RTCE demands exact symbol-level fidelity.

\paragraph{Tokenization is not the bottleneck.}
Tokenization is not the root cause of these failures. First, Llama and the Qwen family use fundamentally different tokenization regimes (e.g., Llama fuses roughly 3 binary digits per token), yet accuracy remains consistently low across both. This rules out a specific tokenizer's behavior as the explanation. Second, QwQ and Qwen2.5-Coder (both 32B) share the exact same parameter count and tokenizer. Despite this identical representation, QwQ achieves a 15.71 Avg. AE score compared to Qwen2.5-Coder's 8.45. This 1.86$\times$ difference is attributable to reasoning-focused training, suggesting the bottleneck is a lack of logical reasoning rather than tokenization.

\subsection{RQ2: Multi-turn revision}
We apply a multi-turn self-reflection approach, in the spirit of~\citep{madaan2023selfrefine,asai2024selfrag}, that refines model answers through structured cycles of critique and revision without regenerating from scratch. Starting from the zero-shot prediction as the initial draft, we run two revision rounds, each comprising two phases. In the \textbf{critique phase}, the model reviews its own current draft and produces a structured assessment: a list of correctness and formatting findings, a suggested fix plan, and an explicit \texttt{VERDICT: KEEP} or \texttt{VERDICT: REVISE} decision. In the \textbf{revision phase}, triggered only when the verdict is \texttt{REVISE}, a separate editor role rewrites the draft strictly according to the critique feedback, without access to the ground truth answer. Before each scoring step, answers are standardised into a canonical JSON format \verb|{"output": <value>}| to decouple formatting errors from genuine reasoning failures. To ensure all prompts fit within the model's context window, the system estimates token usage and applies context budgeting, trimming inputs and adjusting generation lengths as needed. Iteration terminates early as soon as an exact match with the ground truth is produced. If no exact match is achieved after all rounds, fallback policies are applied: the final draft is retained as is, optionally annotated to indicate the lack of confirmed correctness, or replaced with a neutral placeholder that signals prediction uncertainty without revealing the ground truth. We show the exact working configuration in Appendix~\ref{sec:multi-turn-revision-config}.

Figure~\ref{fig:multi-turn-revision-ae} shows performance across revision rounds on AE. The first revision round yields measurable gains, particularly on inversion tasks (\(z \xrightarrow{\invenc} x'\), \(x \xrightarrow{\invdec} z\)), which suffer disproportionately from shallow reasoning errors that a structured critique can surface. However, subsequent rounds exhibit sharply diminishing returns, with performance plateauing well below the ceiling. This rapid saturation reveals a fundamental limitation of unaided self-critique: while it can recover from minor formatting and surface-level reasoning errors, it cannot resolve deeper failures such as systematic state-tracking errors or incorrect inversion logic. These require external execution feedback or stronger compositional reasoning abilities~\citep{Ni_Xiao_Meng_Li_Zheng_Liang_2025,kohli-etal-2025-groundcocoa}, neither of which self-reflection alone provides.

\begin{figure}
    \centering
    \includegraphics[width=1.\linewidth]{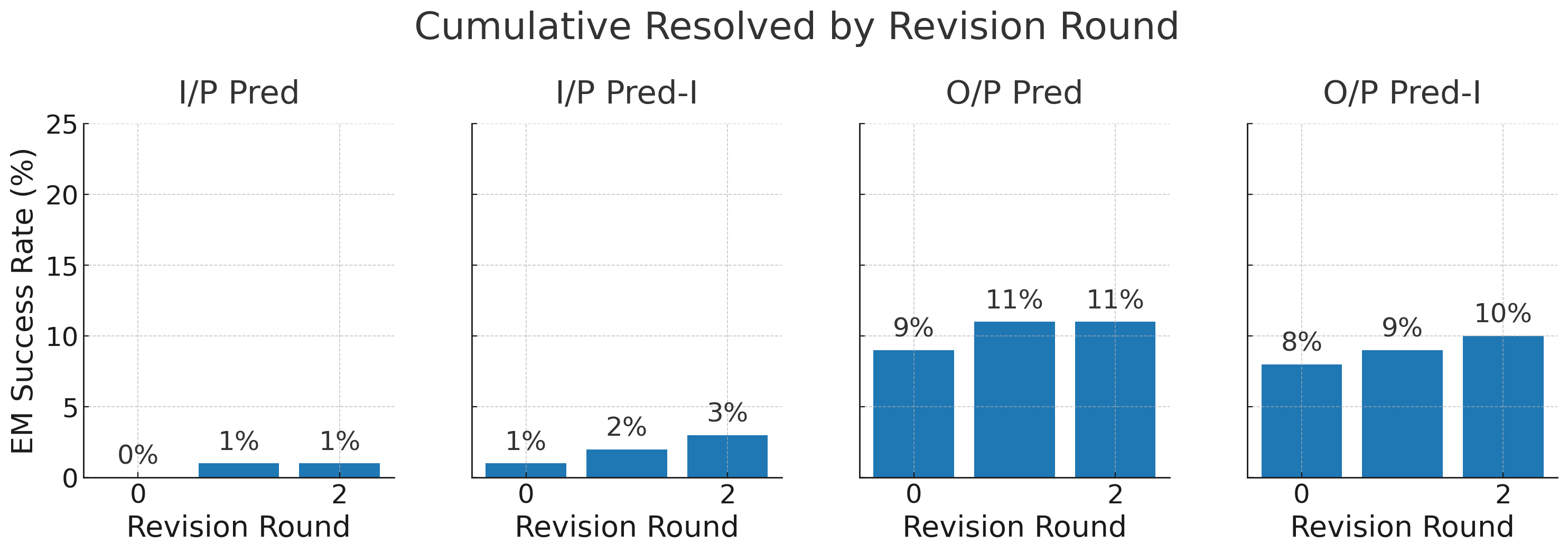}
    \caption{Multi-turn revision for AE on a subset of the dataset. Overall, tasks requiring inversion show lower initial accuracy but larger gains with revision across the two rounds.}
    \label{fig:multi-turn-revision-ae}
\end{figure}

\subsection{RQ3: Fine-tuning}
Our SFT pipeline proceeds in five stages: execution trace generation, trace filtering, natural language translation, supervised data construction, and LoRA fine-tuning.

\paragraph{Trace generation.}
We instrument each reference solution by injecting a \texttt{@snoop} decorator at a depth equal to the number of helper functions defined in the file, enabling fine-grained step-by-step variable tracing. The instrumented script is executed in an isolated subprocess (new process group) with a strict five-second wall-clock timeout; upon expiry, the process tree is terminated via \texttt{SIGTERM}, followed by \texttt{SIGKILL}. Execution traces are captured from \texttt{stderr} and must terminate with the sentinel line \texttt{<<< Return value from main\_solution: <value>}, which simultaneously verifies successful completion and records the ground-truth output.

\paragraph{Trace filtering.}
A trace is accepted only if it (i) contains no runtime error markers, (ii) ends with the sentinel line as its final entry, and (iii) does not exceed 3,000 lines. Applied to the 250-input RTCE pool, this yields 87--98 validated examples per algorithm (87 for RLE, 98 for AE), each associating a problem description and input with a clean, deterministic execution trace.

\paragraph{Natural language translation.}
Validated traces are translated into step-by-step natural-language reasoning using Qwen3-32B, deployed offline via vLLM with sampling parameters: temperature 0.6, top-$p$ 0.95, top-$k$ 20, and a maximum of 8,192 output tokens. Memory addresses are stripped from traces before prompting. The model is instructed to reproduce the exact intermediate values visible in the trace, while writing as if reasoning through the code independently, without citing the trace as a source. The ground-truth answer is parsed from the sentinel line to serve as the evaluation label. Prompts exceeding 32,768 characters are discarded prior to generation.

\paragraph{SFT data construction.}
Each translated example is cast as a two-turn chat instance aligned with the QwQ-32B instruction template: the user message contains the problem description, reference solution code, and input; the assistant message contains the natural-language reasoning chain. Sequences that exceed 32,768 characters after applying the chat template are filtered out, yielding a compact, high-quality supervised fine-tuning dataset.

\paragraph{LoRA fine-tuning.}
We fine-tune QwQ-32B using Low-Rank Adaptation (LoRA; \citealt{hu2022lora}) with DeepSpeed ZeRO-3 across four GPUs. Rank-8 adapters are applied to all linear modules (\texttt{lora\_target: all}). Training runs for three epochs with a per-device batch size of 1 and gradient accumulation of 8 (effective batch size 32), a peak learning rate of $1{\times}10^{-4}$ under a cosine schedule with 10\% linear warmup, a sequence cutoff of 2,048 tokens, and BF16 mixed precision with SDPA flash attention~\citep{dao2022flashattention}. One adapter checkpoint is produced per algorithm. Full hyperparameter details are given in Table~\ref{tab:ft-hyperparams} (Appendix~\ref{sec:fine-tune-hparams}).

Table~\ref{tab:pass5_by_algo_temp_pct_2dp} reports fine-tuning results for QwQ-32B, the strongest zero-shot model. SFT substantially lifts forward task performance on RLE (80.0\%), LZW (87.5\%), and AE (78.6\%), yet inverse task gains are uneven. RLE holds up well (76.5\%), while AE collapses to 30.8\% and Huffman encoding remains at 0\% despite decoder improvement (35.0\%). This asymmetry reveals that exposure to correct forward-execution steps is not sufficient: the adapters overfit to the surface form of trace-derived reasoning chains without internalising the bijective state-transition structure required for inversion.

This confirms our central diagnostic argument: We hypothesise the bottleneck is not exposure to correct reasoning steps but the model's inability to generalise the algorithm's bijective invariants to unseen inputs. Likewise, the rapid saturation of self-reflection after the first revision round (RQ2) indicates that the model lacks the capacity for deep semantic repair rather than merely surface-level error correction. Together, both results converge on the same conclusion: current LLMs fail on RTCE because they cannot maintain and invert evolving algorithmic state~\citep{dziri2023faith}, a limitation that neither trace-supervised fine-tuning nor iterative self-critique can overcome.

\begin{table}[htbp]
\centering
\resizebox{1.\linewidth}{!}{%
\begin{tabular}{lccccc}
\toprule
\textbf{algo} & \textbf{temp} &
\textbf{I/P Pred} &
\textbf{I/P Pred-I} &
\textbf{O/P Pred} &
\textbf{O/P Pred-I} \\
\midrule
\multirow{2}{*}{ae}      & 0.2 & 30.77 & 23.08 & 78.57 & 84.62 \\
                         & 0.8 & 15.00 & 20.00 & 70.00 & 84.21 \\
\midrule
\multirow{2}{*}{huffman} & 0.2 & 35.00 & 50.00 & 0.00  & 0.00  \\
                         & 0.8 & 36.36 & 50.00 & 0.00  & 0.00  \\
\midrule
\multirow{2}{*}{lzw}     & 0.2 & 62.50 & 62.50 & 87.50 & 87.50 \\
                         & 0.8 & 77.42 & 75.00 & 78.13 & 80.65 \\
\midrule
\multirow{2}{*}{rle}     & 0.2 & 76.47 & 86.00 & 80.00 & 86.00 \\
                         & 0.8 & 80.65 & 86.89 & 80.33 & 86.89 \\
\bottomrule
\end{tabular}
}
\caption{Pass@5 after fine-tuning Qwen/QwQ-32B with LoRA across two different temperature settings. Taken together, models get better at one direction or the inversion variant, yet they still fail to maintain high accuracy simultaneously across all four tasks.}
\label{tab:pass5_by_algo_temp_pct_2dp}
\end{table}

\section{Conclusion}
We introduce RTCE to assess round-trip consistency in Code-LLMs, highlighting a new  test for model robustness. Achieving round-trip bijection is a stringent objective, requiring to internalise deep semantic and structural code properties, including variable dependencies, control flow, type constraints, and program invariants. When tasked with both encoding (forward execution) and decoding (inverse reconstruction), LLMs consistently display systematic yet non-obvious inconsistencies. These manifest as information loss, the introduction of hallucinated details, or violations of semantic constraints, preventing exact input recovery and expose brittle, non-compositional reasoning. LLMs, even when assisted with advanced prompts, fine-tuning, or self-reflection, do not fully demonstrate the internal semantic coherence required for trustworthy, compositional code reasoning. This opens a rich landscape for future work that could span new architectures, training objectives, and evaluation methodologies aimed at building models capable of reliable  code reasoning.

\section*{Limitations}
In our analysis, we predominantly use open-source LLMs, as they offer advantages such as ease of fine-tuning, self-reflection, and iterative reasoning capabilities, which are often limited or inaccessible in closed-source models. Additionally, the design of RTCE centres on the compression-decompression paradigm, which may overlook other forms of invertibility, such as refactoring, decompilation, or symbolic manipulation, that are equally important for code inversion. Within the RTCE framework, the current implementation focuses exclusively on Python, which may limit the generalisability of insights into model capabilities across other programming languages, though extending to other languages is straightforward. While 1,000 exact-match instances are sufficient to expose fundamental inversion failures, the small pool limits fine-grained per-category statistical reliability and may underestimate the variance in difficulty across input distributions. Finally, the evaluation framework is static and execution-free, thus unable to assess how models handle runtime phenomena such as side effects, concurrency, exceptions, or external dependencies, which are common challenges in code correctness and invertibility.

\section*{Ethical Considerations}
RTCE is built from synthetic inputs and deterministic reference implementations; it involves no human subjects, crowdsourced annotations, or personally identifiable data. We evaluate only publicly released models, making our results fully reproducible without access to proprietary APIs. RTCE probes a specific reasoning ability: whether LLMs can faithfully execute and invert lossless compression algorithms. This is a diagnostic tool for understanding model limitations, not a system that produces harmful outputs or enables misuse. Large-scale inference and fine-tuning carry a computational cost, which we mitigated by batching requests and reusing completed inference files. The dataset, code, and model outputs are released under permissive licences to support reproducibility.
 
\section*{Acknowledgements}
We thank the reviewers and the area chair for their useful comments. NM was supported by the UKRI Centre for Doctoral Training (CDT) in Natural Language Processing through the UKRI grant (EP/S022481/1). The authors acknowledge the use of resources provided by the Isambard-AI National AI Research Resource (AIRR; \citealt{mcintoshsmith2024isambardaileadershipclasssupercomputer}). Isambard-AI is operated by the University of Bristol and is funded by the UK Government's Department for Science, Innovation and Technology (DSIT) via UK Research and Innovation; and the Science and Technology Facilities Council [ST/AIRR/I-A-I/1023].
AV was supported by the ``UNREAL: Unified Reasoning Layer for Trustworthy ML'' project (EP/Y023838/1) selected by the ERC and funded by UKRI EPSRC.

\bibliography{custom,anthology-1,anthology-2}

\clearpage

\appendix
\section{Concrete task example}
\label{sec:task-example}
Figure~\ref{fig:concrete-task-example} illustrates the four prediction tasks applied to a single RLE input, showing both the forward (encoding) and backward (decoding with inversion) directions through the compression pipeline.
\begin{figure}[ht]
\centering
\begin{tcolorbox}[
  enhanced,
  boxrule=0.5pt,
  left=4pt,right=4pt,top=4pt,bottom=4pt,
  sharp corners,
  title={Concrete Task Example},
  fonttitle=\bfseries\small
]
\footnotesize
\noindent
Consider a simple compression transformation using run-length encoding (RLE), where repeated characters are encoded as character-count pairs:
\[
\enc(x)=\texttt{RLE\_compress}(x), 
\]
\[
\dec(z)=\texttt{RLE\_decompress}(z).
\]
\noindent
Given the input string $x=\texttt{"aaabbbcc"}$ (8 characters), the encoder produces:
\[
z=\enc(x)=\texttt{"a3b3c2"} \quad \text{(6 characters)}.
\]
\noindent
The four evaluation settings are instantiated as follows:
\begin{itemize}[noitemsep,parsep=0pt,leftmargin=*]
  \item \textbf{Output Prediction (forward).}
  Given $(x,\enc)$, the LLM simulates the encoder and predicts:
  \[
  \hat{z}=\enc(x)=\texttt{"a3b3c2"}.
  \]
  
  \item \textbf{Input Prediction with Inversion.}
  Given $(z,\enc)$, the LLM must mentally invert $\enc$ to act as the decoder and predict:
  \[
  \hat{x}'=\enc^{-1}(z)=\dec(z)=\texttt{"aaabbbcc"}.
  \]
  
  \item \textbf{Output Prediction with Inversion.}
  Given $(x,\dec)$, the LLM must mentally invert $\dec$ to act as the encoder and predict:
  \[
  \hat{z}=\dec^{-1}(x)=\enc(x)=\texttt{"a3b3c2"}.
  \]
  
  \item \textbf{Input Prediction (forward).}
  Given $(z,\dec)$, the LLM simulates the decoder and predicts:
  \[
  \hat{x}'=\dec(z)=\texttt{"aaabbbcc"}.
  \]
\end{itemize}
\noindent
In the two \emph{with inversion} tasks (2 and 3), the LLM must mentally invert the provided function to solve the task correctly, even though the inverse function is never explicitly given. This tests whether the model understands the bidirectional relationship between encoding and decoding.
\end{tcolorbox}
\caption{A concrete task example outlining the workflow across the four prediction tasks using run-length encoding.}
\label{fig:concrete-task-example}
\end{figure}

\section{Compression algorithms}
\label{sec:compression-algo}

\begin{lstlisting}[style=py-light, caption={LZW encoding reference.}, label={lst:lzw-enc}]
def main_solution(uncompressed):
    dict_size = 256
    dictionary = {chr(i): i for i in range(dict_size)}
    w = ""
    result = []
    for c in uncompressed:
        wc = w + c
        if wc in dictionary:
            w = wc
        else:
            result.append(dictionary[w])
            dictionary[wc] = dict_size
            dict_size += 1
            w = c
    if w:
        result.append(dictionary[w])
    return result
\end{lstlisting}

\begin{lstlisting}[style=py-light, caption={LZW decoding reference.}, label={lst:lzw-dec}]
def main_solution(compressed):
    dict_size = 256
    dictionary = {i: chr(i) for i in range(dict_size)}
    result = []
    w = chr(compressed.pop(0))
    result.append(w)
    for k in compressed:
        if k in dictionary:
            entry = dictionary[k]
        elif k == dict_size:
            entry = w + w[0]
        else:
            raise ValueError("Bad compressed k: %
        result.append(entry)
        dictionary[dict_size] = w + entry[0]
        dict_size += 1
        w = entry
    return "".join(result)
\end{lstlisting}

\begin{lstlisting}[style=py-light, caption={AE encoding reference (using per-item freq).}, label={lst:ae-enc}]
def main_solution(uncompressed):
    freq = FREQ_DICT  # injected per item, includes 'EOF'
    total = sum(freq.values())
    symbols = sorted(freq.keys())
    cum_counts, running = {}, 0
    for sym in symbols:
        cum_counts[sym] = running
        running += freq[sym]
    low, high = 0.0, 1.0
    for c in list(uncompressed) + ['EOF']:
        width = high - low
        high = low + width * (cum_counts[c] + freq[c]) / total
        low  = low + width *  cum_counts[c] / total
    return (low + high) / 2
\end{lstlisting}

\begin{lstlisting}[style=py-light, caption={AE decoding reference (using per-item freq).}, label={lst:ae-dec}]
def main_solution(compressed):
    freq = FREQ_DICT  # injected per item, includes 'EOF'
    total = sum(freq.values())
    symbols = sorted(freq.keys())
    cum_counts, running = {}, 0
    for s in symbols:
        cum_counts[s] = running
        running += freq[s]
    low, high = 0.0, 1.0
    result = []
    while True:
        width = high - low
        scaled = (compressed - low) / width * total
        for s in symbols:
            if cum_counts[s] <= scaled < cum_counts[s] + freq[s]:
                symbol = s
                break
        if symbol == 'EOF':
            break
        result.append(symbol)
        high = low + width * (cum_counts[symbol] + freq[symbol]) / total
        low  = low + width *  cum_counts[symbol] / total
    return ''.join(result)
\end{lstlisting}

\begin{lstlisting}[style=py-light, caption={RLE encoding reference.}, label={lst:rle-enc}]
def main_solution(uncompressed):
    if not uncompressed:
        return []
    result = []
    prev_char = uncompressed[0]
    count = 1
    for c in uncompressed[1:]:
        if c == prev_char:
            count += 1
        else:
            result.append((prev_char, count))
            prev_char = c
            count = 1
    result.append((prev_char, count))
    return result
\end{lstlisting}

\begin{lstlisting}[style=py-light, caption={RLE decoding reference.}, label={lst:rle-dec}]
def main_solution(compressed):
    result = []
    for char, count in compressed:
        result.append(char * count)
    return ''.join(result)
\end{lstlisting}

\begin{lstlisting}[style=py-light, caption={Huffman encoding reference.}, label={lst:huff-enc}]
def main_solution(uncompressed):
    from collections import namedtuple
    import heapq
    freq = FREQ_DICT  # injected per item
    Node = namedtuple('Node', ['freq','symbol','left','right'])
    Node.__lt__ = lambda a,b: a.freq < b.freq
    heap = [Node(f, sym, None, None) for sym, f in freq.items()]
    heapq.heapify(heap)
    while len(heap) > 1:
        l = heapq.heappop(heap); r = heapq.heappop(heap)
        heapq.heappush(heap, Node(l.freq + r.freq, None, l, r))
    codebook = {}
    def walk(node, prefix):
        if node.symbol is not None:
            codebook[node.symbol] = prefix or '0'
        else:
            walk(node.left, prefix + '0')
            walk(node.right, prefix + '1')
    walk(heap[0], '')
    bitstr = ''.join(codebook[c] for c in uncompressed)
    padding = (-len(bitstr)) %
    bitstr += '0' * padding
    encoded_bytes = [int(bitstr[i:i+8], 2) for i in range(0, len(bitstr), 8)]
    return encoded_bytes, codebook, padding
\end{lstlisting}

\begin{lstlisting}[style=py-light, caption={Huffman decoding reference.}, label={lst:huff-dec}]
def main_solution(compressed):
    from collections import namedtuple
    import heapq
    encoded_bytes, codebook, padding = compressed
    freq = FREQ_DICT  # same dict used to rebuild the tree
    Node = namedtuple('Node', ['freq','symbol','left','right'])
    Node.__lt__ = lambda a,b: a.freq < b.freq
    heap = [Node(f, sym, None, None) for sym, f in freq.items()]
    heapq.heapify(heap)
    while len(heap) > 1:
        l = heapq.heappop(heap); r = heapq.heappop(heap)
        heapq.heappush(heap, Node(l.freq + r.freq, None, l, r))
    root = heap[0]
    bitstr = ''.join(f'{b:08b}' for b in encoded_bytes)
    if padding:
        bitstr = bitstr[:-padding]
    result, node = [], root
    for bit in bitstr:
        node = node.left if bit == '0' else node.right
        if node.symbol is not None:
            result.append(node.symbol)
            node = root
    return ''.join(result)
\end{lstlisting}

\section{Model inference details}
\label{sec:model-inference}
All experiments employed the \texttt{vLLM} inference engine. Decoding was performed with temperatures $\{0.2, 0.8\}$ and nucleus sampling ($p=0.95$), producing up to $16{,}384$ tokens per completion. Unless specified otherwise, each prompt generated $n=5$ completions. Models were initially loaded in \texttt{bfloat16} precision with the configured tensor parallelism (\texttt{tp\_size}); on GPU memory exhaustion, inference fell back to 4-bit NF4 quantisation via \texttt{bitsandbytes} with double quantisation and \texttt{float16} compute. A fixed random seed ($42$) ensured reproducibility during both data curation and model inference. The stop sequence \texttt{[\,/ANSWER]} was enforced with inclusion in the output. For robustness, execution time was logged, and each request was retried up to two times upon failure with a $5$\,s backoff. The outputs were stored in JSONL format, containing model metadata (name, size, category), inference parameters, and generated content.

Our experiments were conducted using four NVIDIA GH200 GPUs, each featuring 120GB of memory. For each model and temperature setting, we estimate an average inference time of approximately 14 hours to produce predictions of 5000 input samples.

\section{Fine-tuning hyperparameters}
\label{sec:fine-tune-hparams}

Table~\ref{tab:ft-hyperparams} shows the configured values for the set of hyperparameters used in the fine-tuning experiments.
\begin{table}[htbp]
\centering
\resizebox{0.7\linewidth}{!}{%
\begin{tabular}{ll}
\toprule
\textbf{Hyperparameter} & \textbf{Value} \\
\midrule
Base model & Qwen/QwQ-32B \\
Fine-tuning type & LoRA \\
LoRA rank & 8 \\
LoRA target & all \\
Stage & SFT \\
Epochs & 3.0 \\
Learning rate & $1 \times 10^{-4}$ \\
Batch size (per device) & 1 \\
Gradient accumulation & 8 \\
Cutoff length & 2048 \\
Precision & FP16 \\
\bottomrule
\end{tabular}
}
\caption{Key fine-tuning hyperparameters for Qwen/QwQ-32B with LoRA.}
\label{tab:ft-hyperparams}
\end{table}

\section{Multi-turn revision configurations}
\label{sec:multi-turn-revision-config}
\paragraph{Environment and hardware.} 
Experiments were run under SLURM using a Singularity container 
\texttt{e4s-cuda90-aarch64-25.06.sif}. 
We used Python~3.10 in a Conda environment created with 
\texttt{requirements.txt}. 
Jobs requested 4~GPUs, 32~CPUs, and 128~GB RAM, though each run was launched with 
\texttt{--num\_gpus 1}. 
Offline mode was enabled (\texttt{--hf\_offline}), with the Hugging Face cache stored at 
\texttt{\textasciitilde/.cache/huggingface/hub}. 

\paragraph{Models.} 
All experiments used the \texttt{Qwen/QwQ-32B} model. 
We applied RoPE scaling with the following JSON configuration:
\begin{verbatim}
{"rope_type":"yarn","factor":16.0,
 "original_max_position_embeddings":8192}
\end{verbatim}

\paragraph{Data.} 
Algorithms evaluated were: \texttt{ae}, \texttt{lzw}, \texttt{rle}, and \texttt{huffman}. 
For each, the input file was of the form:
\begin{verbatim}
processed_datasets_test/${ALGO}/
  ${MODEL}_temp_${TEMP}_n5_verified.jsonl
\end{verbatim}
The ground truth field was \texttt{res.actual}. 
Temperature was parsed from the filename (e.g.\ \texttt{\_temp\_0.2\_} $\to$ 0.2).

\paragraph{Core settings.} 
We used 2 reflection rounds (\texttt{--reflection\_rounds 2}), 
critique style B (\texttt{--critique\_style B}), and early stopping on EM 
(\texttt{--gt\_stop\_on em}). 
The mismatch policy was \texttt{--on\_mismatch annotate}, which keeps the draft 
and attaches a status note (no ground truth leak). 
Additional context and generation settings were:
\begin{verbatim}
--max_model_len 65536
--model_ctx 8192
--gen_tokens 512
--max_tokens 1024
--safety_margin 64
--truncate_hard_chars 16000
--chars_per_token 1.5
--gpu_memory_utilization 0.9
\end{verbatim}

\paragraph{Answer schema and canonicalization.} 
Outputs were required in the form \verb|{"output": <value>}|, optionally wrapped in 
\verb|[ANSWER]...[/ANSWER]| tags (\texttt{--force\_answer\_tags}). 
The canonicalizer strips code fences, scans up to 20k characters for the first 
valid JSON object, and flattens nested forms such as 
\verb|{"output":{"return":x}}| $\to$ \verb|x|. 
If no valid JSON is found, a format-only repair prompt rewrites the output 
into the required schema.

\paragraph{Scoring.} 
Exact match (EM) was computed as follows: numbers are compared using 
\texttt{isclose} with relative and absolute tolerance $10^{-3}$; strings are 
case-folded, trimmed, and unquoted if necessary.

\paragraph{Prompts (verbatim).}
\textbf{Critique-B (system):}
\begin{verbatim}
You are a structured reviewer. 
Provide actionable findings 
and a fix plan. Do not reveal or 
approximate any expected value.
\end{verbatim}

\textbf{Critique-B (user):}
\begin{verbatim}
Conversation:
{conversation}

Draft answer:
{draft}

Write sections 'Findings:' and 'Fix:' 
in bullet points.
Do not include or infer the expected value.
End with VERDICT: KEEP or VERDICT: REVISE.
\end{verbatim}

\textbf{Revision (no-leak, system):}
\begin{verbatim}
You are a careful editor. Revise the draft 
strictly according to the feedback. 
Do not include analysis. 
Provide only the improved final answer.
\end{verbatim}

\textbf{Revision (no-leak, user):}
\begin{verbatim}
Conversation:
{conversation}

Draft answer:
{draft}

Feedback:
{feedback}

Now produce the corrected final answer only.
\end{verbatim}

\textbf{Format-repair (system):}
\begin{verbatim}
You must output ONLY:
[ANSWER]
{"output": <value>}
[/ANSWER]
No other text, code, or explanations.
\end{verbatim}

\textbf{Format-repair (user):}
\begin{verbatim}
Rewrite into the required format. 
If it already contains the 
needed value, keep it.
------
{raw_text}
------
\end{verbatim}

\section{Prompt templates}
\label{sec:prompt-templates}

We show the Prompt Templates employed in our experiments, capturing the structural variations in how tasks were framed for the models.~\cref{fig:ip-prompt,fig:op-prompt} illustrate the prompt templates for input prediction and output prediction, respectively. In addition,~\cref{fig:ip-i-prompt,fig:op-i-prompt} present the corresponding prompt templates that incorporate inversion, demonstrating how the task formulation changes under this modification.

\begin{figure*}[htbp]
\centering
\begin{minipage}{\linewidth} %
\begin{tcolorbox}[
  enhanced,
  boxrule=0.5pt,
  left=4pt,right=4pt,top=4pt,bottom=4pt,
  sharp corners,
  title={Input prediction prompt template},
  fonttitle=\bfseries\small
]
{\small
\textbf{System.} You are a helpful programming assistant designed to execute code. You must verify your output via a round-trip check and self-correct before returning the final JSON.

\medskip
\textbf{User.} You are given a Python function and an input. Return a \emph{literal} output (no unsimplified expressions, no function calls), enclosed within \texttt{[ANSWER]} and \texttt{[/ANSWER]} tags. Do \emph{not} include any additional text.

\medskip
The input and output requirements are as follows:

\textbf{Input:}  
\quad \texttt{s} (\texttt{str}): The input string to be duplicated and wrapped.

\textbf{Output:}  
\quad \texttt{return} (\texttt{str}): A string starting with ``b'', followed by two copies of \texttt{s}, and ending with ``a''.

\medskip
Given the following input:

\begin{verbatim}
"hi"
\end{verbatim}

Given the following function:
\begin{verbatim}
[PYTHON]
def main_solution(s):
    s = s + s
    return "b" + s + "a"
[/PYTHON]
\end{verbatim}

Can you predict the output without writing any code? Do not include any explanations, reasoning, or extra text. Put your final answer in the following JSON format:  
\texttt{{"output": <your output>}}, where \texttt{<your output>} must strictly match the output requirement specified above.

\medskip
[THOUGHT]

Let's execute the code step by step:

1. The function \texttt{main\_solution} is defined, which takes a single argument \texttt{s}.  

2. The function is called with the argument \texttt{"hi"}, so within the function, \texttt{s} is initially \texttt{"hi"}.  

3. Inside the function, \texttt{s} is concatenated with itself, so \texttt{s} becomes \texttt{"hihi"}.  

4. The function then returns a new string that starts with \texttt{"b"}, followed by the value of \texttt{s} (which is now \texttt{"hihi"}), and ends with \texttt{"a"}.  

5. The return value of the function is therefore \texttt{"bhihia"}. 

[/THOUGHT]

\medskip
[ANSWER]

\{"output": "bhihia"\}  

[/ANSWER]

\medskip
The input and output requirements are as follows:

\textbf{Input:}  
\quad \texttt{uncompressed} (\texttt{str}): The input string to be compressed.

\textbf{Output:}  
\quad \texttt{return} (list of tuple): A list of \texttt{(char, count)} tuples representing the RLE-compressed string.

\medskip
Given the following output:
\begin{verbatim}
<output>
\end{verbatim}

Given the following function:
\begin{verbatim}
<decoding_function>
\end{verbatim}

Can you predict the output without writing any code? Do not include any explanations, reasoning, or extra text. Put your final answer in the following JSON format:  
\texttt{{"output": <your output>}}, where \texttt{<your output>} must strictly match the specified output requirement.

\medskip
[THOUGHT]
} %
\end{tcolorbox}
\end{minipage}
\caption{Input prediction prompt template for RLE algorithm.}
\label{fig:ip-prompt}
\end{figure*}

\begin{figure*}[htbp]
\centering
\begin{minipage}{\linewidth} %
\begin{tcolorbox}[
  enhanced,
  boxrule=0.5pt,
  left=4pt,right=4pt,top=4pt,bottom=4pt,
  sharp corners,
  title={Input prediction with inversion prompt template},
  fonttitle=\bfseries\small
]
{\small
\textbf{System.} You are a helpful programming assistant designed to execute code. You must verify your output via a round-trip check and self-correct before returning the final JSON.

\medskip
\textbf{User.} You are given a Python function and an input. Return a \emph{literal} output (no unsimplified expressions, no function calls), enclosed within \texttt{[ANSWER]} and \texttt{[/ANSWER]} tags. Do \emph{not} include any additional text.

\medskip
The input and output requirements are as follows:

\textbf{Input:}  
\quad \texttt{s} (\texttt{str}): The input string to be duplicated and wrapped.

\textbf{Output:}  
\quad \texttt{return} (\texttt{str}): A string starting with ``b'', followed by two copies of \texttt{s}, and ending with ``a''.

\medskip
Given the following input:

\begin{verbatim}
"hi"
\end{verbatim}

Given the following function:
\begin{verbatim}
[PYTHON]
def main_solution(s):
    s = s + s
    return "b" + s + "a"
[/PYTHON]
\end{verbatim}

Can you predict the output without writing any code? Do not include any explanations, reasoning, or extra text. Put your final answer in the following JSON format:  
\texttt{{"output": <your output>}}, where \texttt{<your output>} must strictly match the output requirement specified above.

\medskip
[THOUGHT]

Let's execute the code step by step:

1. The function \texttt{main\_solution} is defined, which takes a single argument \texttt{s}.  

2. The function is called with the argument \texttt{"hi"}, so within the function, \texttt{s} is initially \texttt{"hi"}.  

3. Inside the function, \texttt{s} is concatenated with itself, so \texttt{s} becomes \texttt{"hihi"}.  

4. The function then returns a new string that starts with \texttt{"b"}, followed by the value of \texttt{s} (which is now \texttt{"hihi"}), and ends with \texttt{"a"}.  

5. The return value of the function is therefore \texttt{"bhihia"}. 

[/THOUGHT]

\medskip
[ANSWER]

\{"output": "bhihia"\}  

[/ANSWER]

\medskip
The input and output requirements are as follows:

\textbf{Input:}  
\quad \texttt{uncompressed} (\texttt{str}): The input string to be compressed.

\textbf{Output:}  
\quad \texttt{return} (list of tuple): A list of \texttt{(char, count)} tuples representing the RLE-compressed string.

\medskip
Given the following output:
\begin{verbatim}
<output>
\end{verbatim}

Given the following function:
\begin{verbatim}
<encoding_function>
\end{verbatim}

The function \texttt{main\_solution} performs encoding. You must use the inverse logic to implement the decoding function, \texttt{main\_solution\_inverse}, to infer your answer, not run or duplicate it directly.
\medskip

Can you predict the output based on \texttt{main\_solution\_inverse}? Do not include any explanations, reasoning, or extra text. Put your final answer in the following json format: {"output": <your output>}, where <your output> should strictly match the output requirement as specified.

\medskip
[THOUGHT]
} %
\end{tcolorbox}
\end{minipage}
\caption{Input prediction with inversion prompt template for RLE algorithm.}
\label{fig:ip-i-prompt}
\end{figure*}

\begin{figure*}[htbp]
\centering
\begin{minipage}{\linewidth} %
\begin{tcolorbox}[
  enhanced,
  boxrule=0.5pt,
  left=4pt,right=4pt,top=4pt,bottom=4pt,
  sharp corners,
  title={Output prediction prompt template},
  fonttitle=\bfseries\small
]
{\small
\textbf{System.} You are a helpful programming assistant designed to execute code. You must verify your output via a round-trip check and self-correct before returning the final JSON.

\medskip
\textbf{User.} You are given a Python function and an input. Return a \emph{literal} output (no unsimplified expressions, no function calls), enclosed within \texttt{[ANSWER]} and \texttt{[/ANSWER]} tags. Do \emph{not} include any additional text.

\medskip
The input and output requirements are as follows:

\textbf{Input:}  
\quad \texttt{s} (\texttt{str}): The input string to be duplicated and wrapped.

\textbf{Output:}  
\quad \texttt{return} (\texttt{str}): A string starting with ``b'', followed by two copies of \texttt{s}, and ending with ``a''.

\medskip
Given the following input:

\begin{verbatim}
"hi"
\end{verbatim}

Given the following function:
\begin{verbatim}
[PYTHON]
def main_solution(s):
    s = s + s
    return "b" + s + "a"
[/PYTHON]
\end{verbatim}

Can you predict the output without writing any code? Do not include any explanations, reasoning, or extra text. Put your final answer in the following JSON format:  
\texttt{{"output": <your output>}}, where \texttt{<your output>} must strictly match the output requirement specified above.

\medskip
[THOUGHT]

Let's execute the code step by step:

1. The function \texttt{main\_solution} is defined, which takes a single argument \texttt{s}.  

2. The function is called with the argument \texttt{"hi"}, so within the function, \texttt{s} is initially \texttt{"hi"}.  

3. Inside the function, \texttt{s} is concatenated with itself, so \texttt{s} becomes \texttt{"hihi"}.  

4. The function then returns a new string that starts with \texttt{"b"}, followed by the value of \texttt{s} (which is now \texttt{"hihi"}), and ends with \texttt{"a"}.  

5. The return value of the function is therefore \texttt{"bhihia"}. 

[/THOUGHT]

\medskip
[ANSWER]

\{"output": "bhihia"\}  

[/ANSWER]

\medskip
The input and output requirements are as follows:

\textbf{Input:}  
\quad \texttt{uncompressed} (\texttt{str}): The input string to be compressed.

\textbf{Output:}  
\quad \texttt{return} (list of tuple): A list of \texttt{(char, count)} tuples representing the RLE-compressed string.

\medskip
Given the following input:
\begin{verbatim}
<input>
\end{verbatim}

Given the following function:
\begin{verbatim}
<encoding_function>
\end{verbatim}

Can you predict the output without writing any code? Do not include any explanations, reasoning, or extra text. Put your final answer in the following JSON format:  
\texttt{{"output": <your output>}}, where \texttt{<your output>} must strictly match the specified output requirement.

\medskip
[THOUGHT]
} %
\end{tcolorbox}
\end{minipage}
\caption{Output prediction prompt template for RLE algorithm.}
\label{fig:op-prompt}
\end{figure*}

\begin{figure*}[htbp]
\centering
\begin{minipage}{\linewidth} %
\begin{tcolorbox}[
  enhanced,
  boxrule=0.5pt,
  left=4pt,right=4pt,top=4pt,bottom=4pt,
  sharp corners,
  title={Output prediction with inversion prompt template},
  fonttitle=\bfseries\small
]
{\small
\textbf{System.} You are a helpful programming assistant designed to execute code. You must verify your output via a round-trip check and self-correct before returning the final JSON.

\medskip
\textbf{User.} You are given a Python function and an input. Return a \emph{literal} output (no unsimplified expressions, no function calls), enclosed within \texttt{[ANSWER]} and \texttt{[/ANSWER]} tags. Do \emph{not} include any additional text.

\medskip
The input and output requirements are as follows:

\textbf{Input:}  
\quad \texttt{s} (\texttt{str}): The input string to be duplicated and wrapped.

\textbf{Output:}  
\quad \texttt{return} (\texttt{str}): A string starting with ``b'', followed by two copies of \texttt{s}, and ending with ``a''.

\medskip
Given the following input:

\begin{verbatim}
"hi"
\end{verbatim}

Given the following function:
\begin{verbatim}
[PYTHON]
def main_solution(s):
    s = s + s
    return "b" + s + "a"
[/PYTHON]
\end{verbatim}

Can you predict the output without writing any code? Do not include any explanations, reasoning, or extra text. Put your final answer in the following JSON format:  
\texttt{{"output": <your output>}}, where \texttt{<your output>} must strictly match the output requirement specified above.

\medskip
[THOUGHT]

Let's execute the code step by step:

1. The function \texttt{main\_solution} is defined, which takes a single argument \texttt{s}.  

2. The function is called with the argument \texttt{"hi"}, so within the function, \texttt{s} is initially \texttt{"hi"}.  

3. Inside the function, \texttt{s} is concatenated with itself, so \texttt{s} becomes \texttt{"hihi"}.  

4. The function then returns a new string that starts with \texttt{"b"}, followed by the value of \texttt{s} (which is now \texttt{"hihi"}), and ends with \texttt{"a"}.  

5. The return value of the function is therefore \texttt{"bhihia"}. 

[/THOUGHT]

\medskip
[ANSWER]

\{"output": "bhihia"\}  

[/ANSWER]

\medskip
The input and output requirements are as follows:

\textbf{Input:}  
\quad \texttt{uncompressed} (\texttt{str}): The input string to be compressed.

\textbf{Output:}  
\quad \texttt{return} (list of tuple): A list of \texttt{(char, count)} tuples representing the RLE-compressed string.

\medskip
Given the following input:
\begin{verbatim}
<input>
\end{verbatim}

Given the following function:
\begin{verbatim}
<decoding_function>
\end{verbatim}

The function \texttt{main\_solution} performs decoding. You must use the inverse logic to implement the encoding function, \texttt{main\_solution\_inverse}, to infer your answer, not run or duplicate it directly.
\medskip

Can you predict the output based on \texttt{main\_solution\_inverse}? Do not include any explanations, reasoning, or extra text. Put your final answer in the following json format: {"output": <your output>}, where <your output> should strictly match the output requirement as specified.

\medskip
[THOUGHT]
} %
\end{tcolorbox}
\end{minipage}
\caption{Output prediction with inversion prompt template for RLE algorithm.}
\label{fig:op-i-prompt}
\end{figure*}

\section{Model URLs}
Table~\ref{tab:model-urls} lists the HuggingFace model URLs for the remaining evaluated models.
\begin{table*}[ht]
\centering
\resizebox{0.9\linewidth}{!}{%
  \begin{tabular}{@{} l r l @{}}
    \toprule
    Model               & Size (B) & HuggingFace URL                                                                 \\
    \midrule
    Yi-Coder-9B-Chat    &  8.80    & \url{https://huggingface.co/01-ai/Yi-Coder-9B-Chat}                    \\
    codegemma-7b-it     &  8.50    & \url{https://huggingface.co/google/codegemma-7b-it}                    \\
    Qwen3-4B            &  4.00    & \url{https://huggingface.co/Qwen/Qwen3-4B}                             \\
    Qwen3-8B            &  8.00    & \url{https://huggingface.co/Qwen/Qwen3-8B}                             \\
    Qwen3-32B           & 32.80    & \url{https://huggingface.co/Qwen/Qwen3-32B}                            \\
    starcoder2-15b-instruct-v0.1 & 15.00 & \url{https://huggingface.co/bigcode/starcoder2-15b-instruct-v0.1}   \\
    CodeLlama-70b-Python-hf      & 70.00 & \url{https://huggingface.co/codellama/CodeLlama-70b-Python-hf}     \\
    CodeLlama-34b-Instruct-hf    & 34.00 & \url{https://huggingface.co/codellama/CodeLlama-34b-Instruct-hf}   \\
    Phi-3-mini-128k-instruct     &  3.80 & \url{https://huggingface.co/microsoft/Phi-3-mini-128k-instruct}    \\
    Phi-3.5-mini-instruct        &  3.80 & \url{https://huggingface.co/microsoft/Phi-3.5-mini-instruct}       \\
    Llama-3.1-8B-Instruct        &  8.03 & \url{https://huggingface.co/meta-llama/Llama-3.1-8B-Instruct}      \\
    Llama-3.1-70B-Instruct       & 70.00 & \url{https://huggingface.co/meta-llama/Llama-3.1-70B-Instruct}     \\
    Llama-3.2-1B-Instruct        &  1.00 & \url{https://huggingface.co/meta-llama/Llama-3.2-1B-Instruct}      \\
    Llama-3.2-3B-Instruct        &  3.00 & \url{https://huggingface.co/meta-llama/Llama-3.2-3B-Instruct}      \\
    phi-2                         &  2.78 & \url{https://huggingface.co/microsoft/phi-2}                       \\
    phi-4                         & 14.70 & \url{https://huggingface.co/microsoft/phi-4}                       \\
    Codestral-22B-v0.1           & 22.20 & \url{https://huggingface.co/mistralai/Codestral-22B-v0.1}          \\
    Mistral-7B-Instruct-v0.3     &  7.24 & \url{https://huggingface.co/mistralai/Mistral-7B-Instruct-v0.3}    \\
    DeepSeek-R1-Distill-Llama-8B &  8.03 & \url{https://huggingface.co/deepseek-ai/DeepSeek-R1-Distill-Llama-8B} \\
    DeepSeek-R1-Distill-Qwen-1.5B&  1.50 & \url{https://huggingface.co/deepseek-ai/DeepSeek-R1-Distill-Qwen-1.5B}\\
    DeepSeek-R1-Distill-Qwen-14B & 14.80 & \url{https://huggingface.co/deepseek-ai/DeepSeek-R1-Distill-Qwen-14B} \\
    DeepSeek-R1-Distill-Qwen-32B & 32.80 & \url{https://huggingface.co/deepseek-ai/DeepSeek-R1-Distill-Qwen-32B} \\
    DeepSeek-R1-Distill-Llama-70B& 70.60 & \url{https://huggingface.co/deepseek-ai/DeepSeek-R1-Distill-Llama-70B}\\
    DeepSeek-R1-0528-Qwen3-8B     &  8.19 & \url{https://huggingface.co/deepseek-ai/DeepSeek-R1-0528-Qwen3-8B} \\
    deepseek-coder-33b-instruct  & 33.30 & \url{https://huggingface.co/deepseek-ai/deepseek-coder-33b-instruct} \\
    Qwen2.5-7B-Instruct          &  7.62 & \url{https://huggingface.co/Qwen/Qwen2.5-7B-Instruct}               \\
    Qwen2.5-72B-Instruct         & 72.00 & \url{https://huggingface.co/Qwen/Qwen2.5-72B-Instruct}              \\
    Qwen2.5-Coder-14B-Instruct   & 14.80 & \url{https://huggingface.co/Qwen/Qwen2.5-Coder-14B-Instruct}        \\
    Qwen2.5-Coder-32B-Instruct   & 32.80 & \url{https://huggingface.co/Qwen/Qwen2.5-Coder-32B-Instruct}        \\
    QwQ-32B                       & 32.80 & \url{https://huggingface.co/Qwen/QwQ-32B}                          \\
    gpt-4o-mini                   &  8.00 & \url{https://huggingface.co/gpt-4o-mini}                           \\
    WizardLM-70B-V1.0             & 70.00 & \url{https://huggingface.co/WizardLMTeam/WizardLM-70B-V1.0}        \\
    \bottomrule
  \end{tabular}%
}
\caption{Model URLs used in our experiments.}
\label{tab:model-urls}
\end{table*}

\section{Pass@5 Radial Plots}
\label{sec:radial-plots}
\begin{figure*}[htbp]
  \centering
  \includegraphics[width=\linewidth]{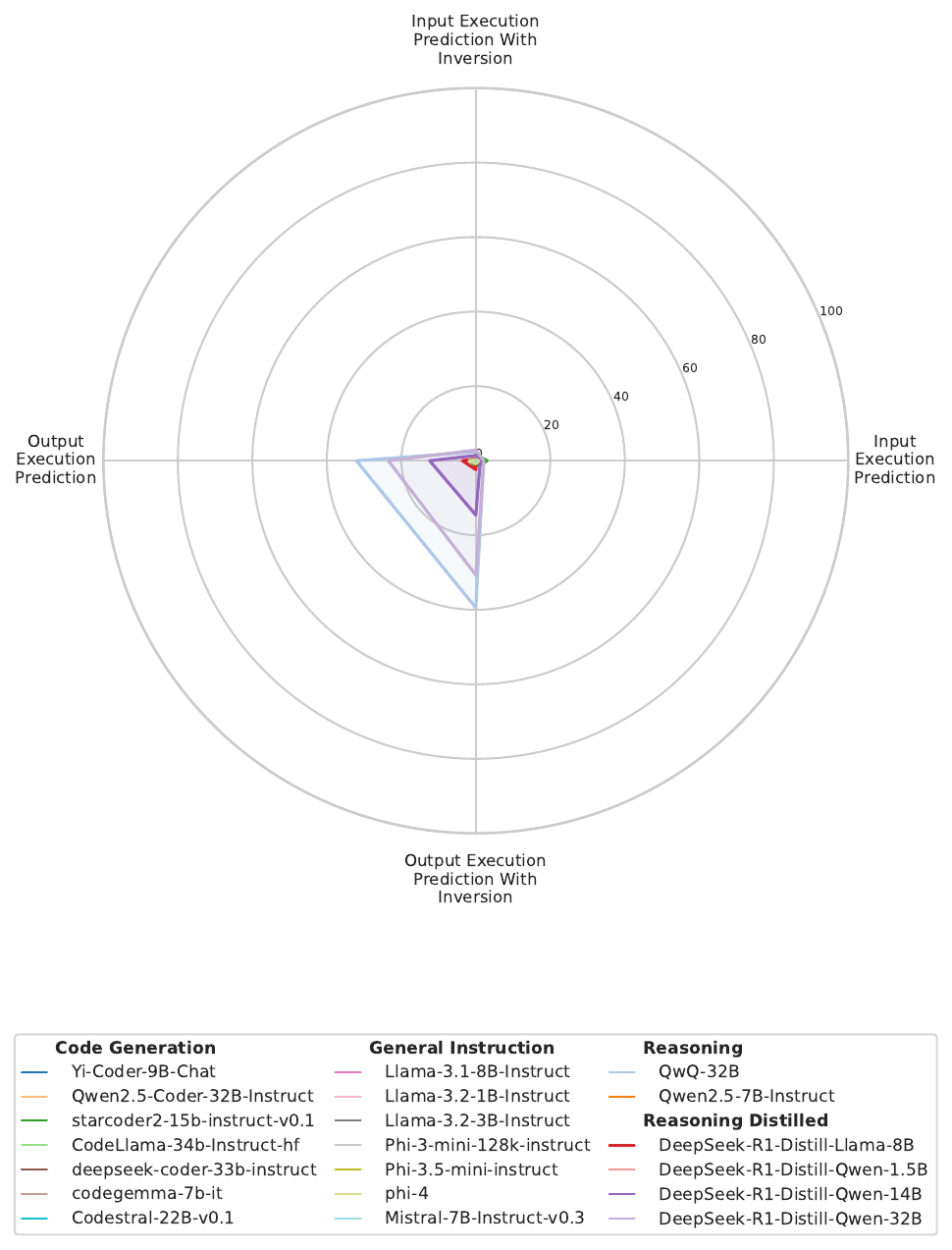}
  \caption{Pass@5 radial plot for AE: models achieve moderate scores on output prediction but collapse on inversion tasks, with larger models showing a more balanced polygon and smaller models nearly flat across all axes.}
  \label{fig:radial-plot-ae}
\end{figure*}

\begin{figure*}[htbp]
  \centering
  \includegraphics[width=\linewidth]{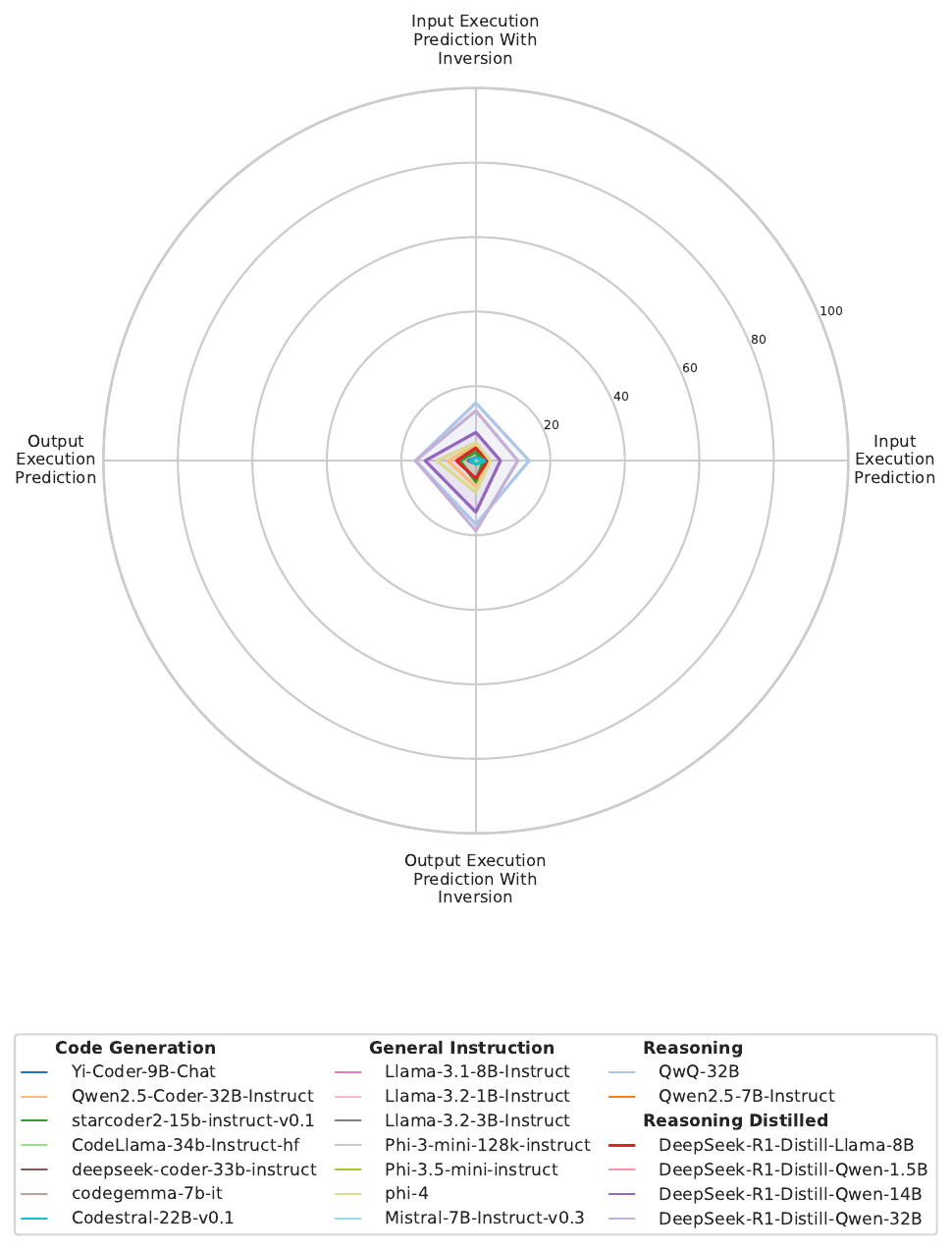}
  \caption{Pass@5 radial plot for LZW: dictionary-based encoding produces a more asymmetric profile than AE, with output prediction axes scoring higher than input prediction axes across nearly all model families.}
  \label{fig:radial-plot-lzw}
\end{figure*}

\begin{figure*}[htbp]
  \centering
  \includegraphics[width=\linewidth]{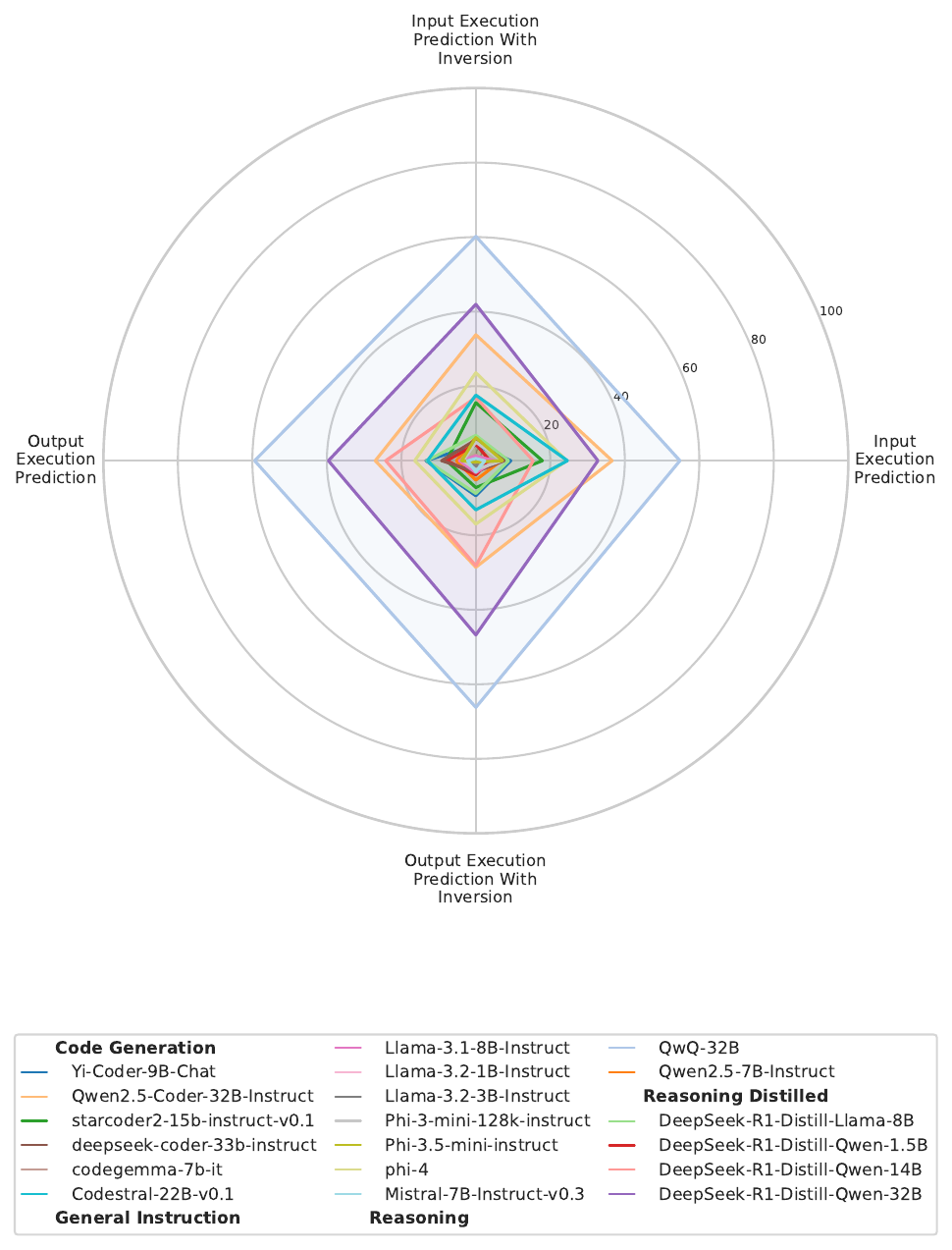}
  \caption{Pass@5 radial plot for RLE: the most filled polygons of all algorithms, confirming RLE is the most tractable; reasoning-focused models (QwQ-32B, DeepSeek-R1) show notably larger and more symmetric shapes.}
  \label{fig:radial-plot-rle}
\end{figure*}

\begin{figure*}[htbp]
  \centering
  \includegraphics[width=\linewidth]{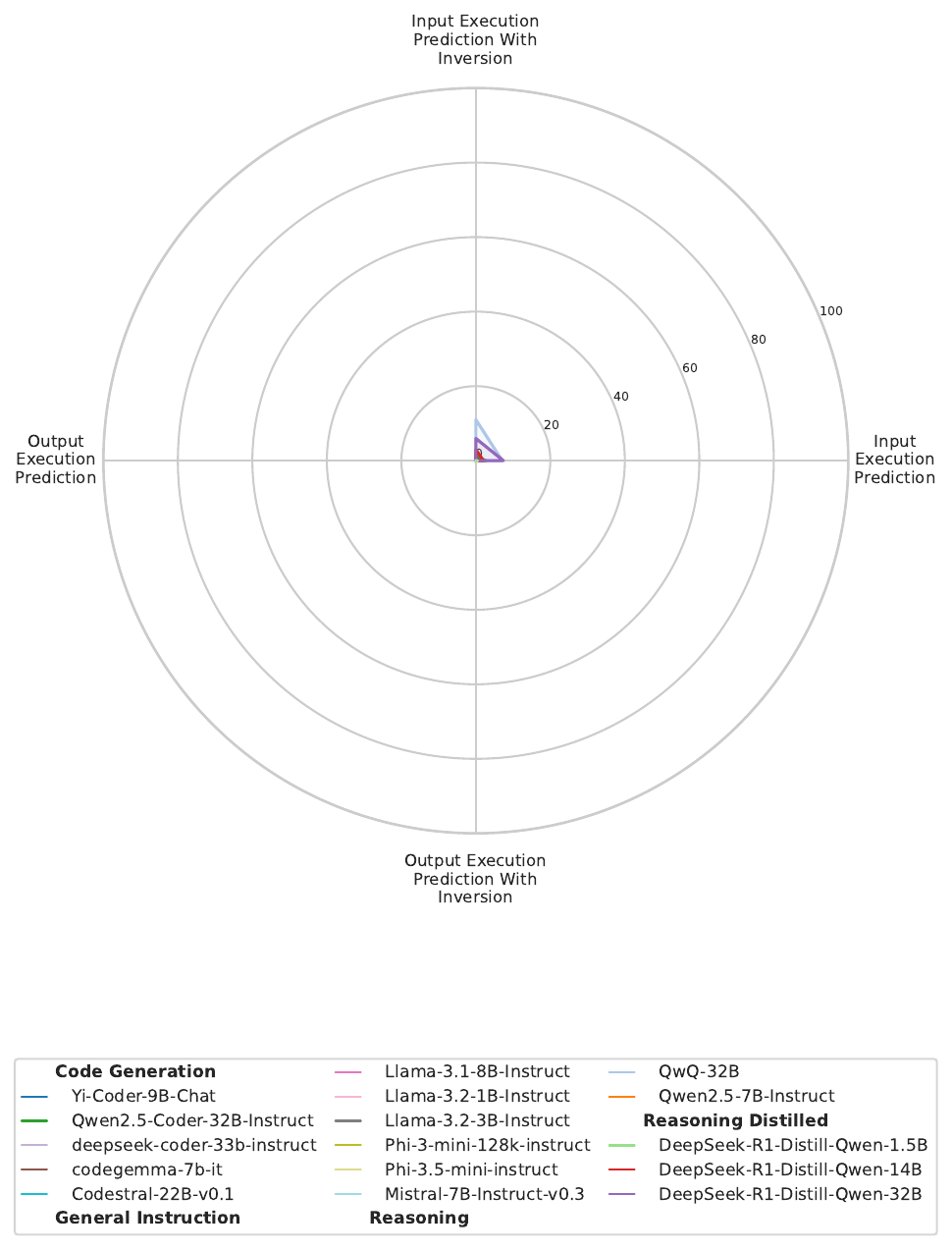}
  \caption{Pass@5 radial plot for Huffman: almost all models collapse to near-zero on every axis, reflecting the extreme difficulty of Huffman encoding; only the largest reasoning models show any measurable area.}
  \label{fig:radial-plot-huffman}
\end{figure*}

Figures~\ref{fig:radial-plot-ae}--\ref{fig:radial-plot-huffman} show per-model Pass@5 radial plots across all four I/O prediction tasks for each compression algorithm. Each axis corresponds to one task, and each line represents a model, allowing direct visual comparison of strengths and weaknesses across tasks and model families.

\section{Input length vs.\ pass@5}
\label{sec:input-len-pass@5}
\begin{figure*}
    \centering
    \includegraphics[width=1\linewidth]{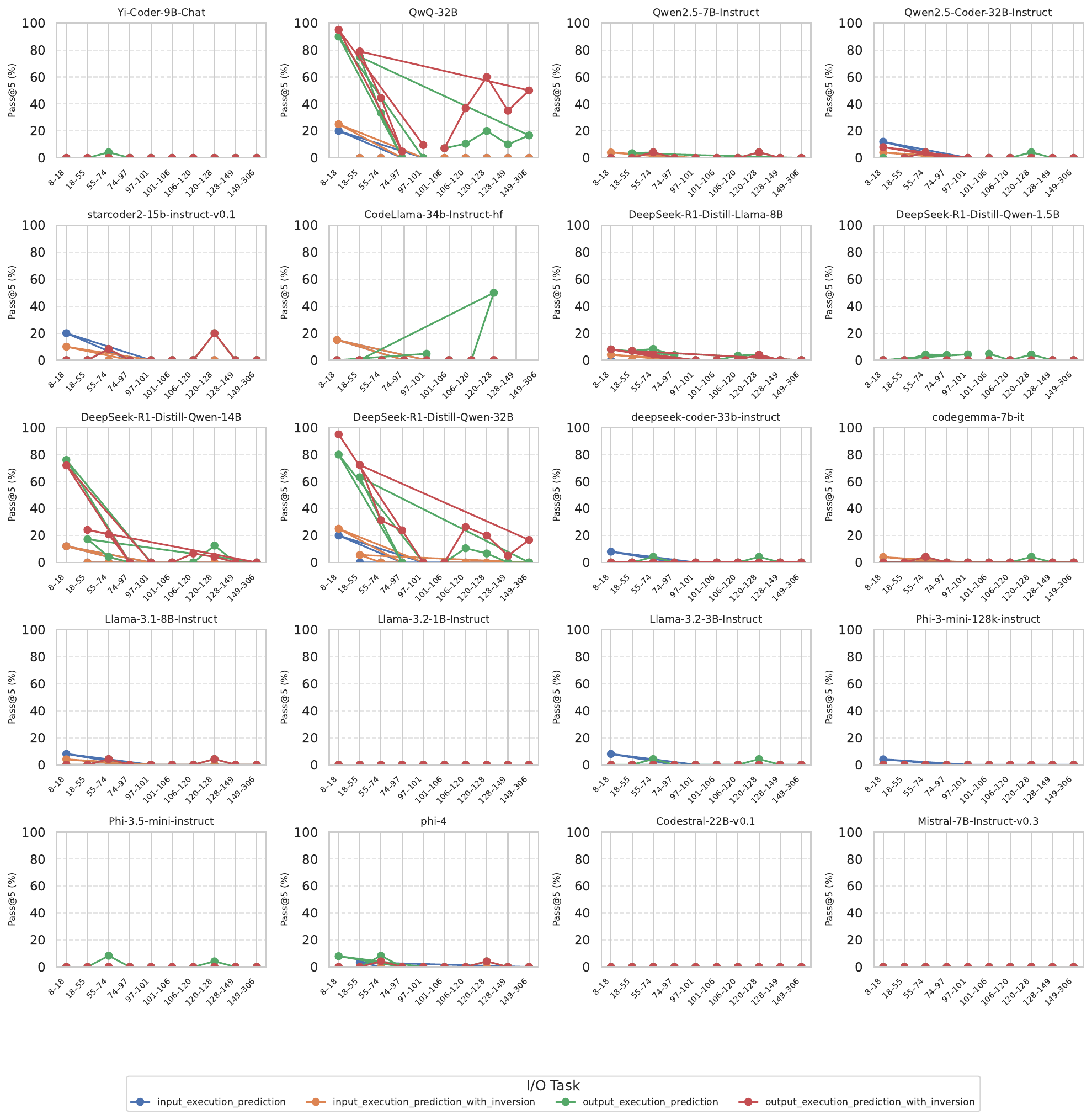}
    \caption{Input length vs.\ Pass@5 for AE: Pass@5 drops sharply beyond short inputs, with nearly all models failing on strings longer than 100 characters.}
    \label{fig:input-len-pass-5-ae}
\end{figure*}

\begin{figure*}
    \centering
    \includegraphics[width=1\linewidth]{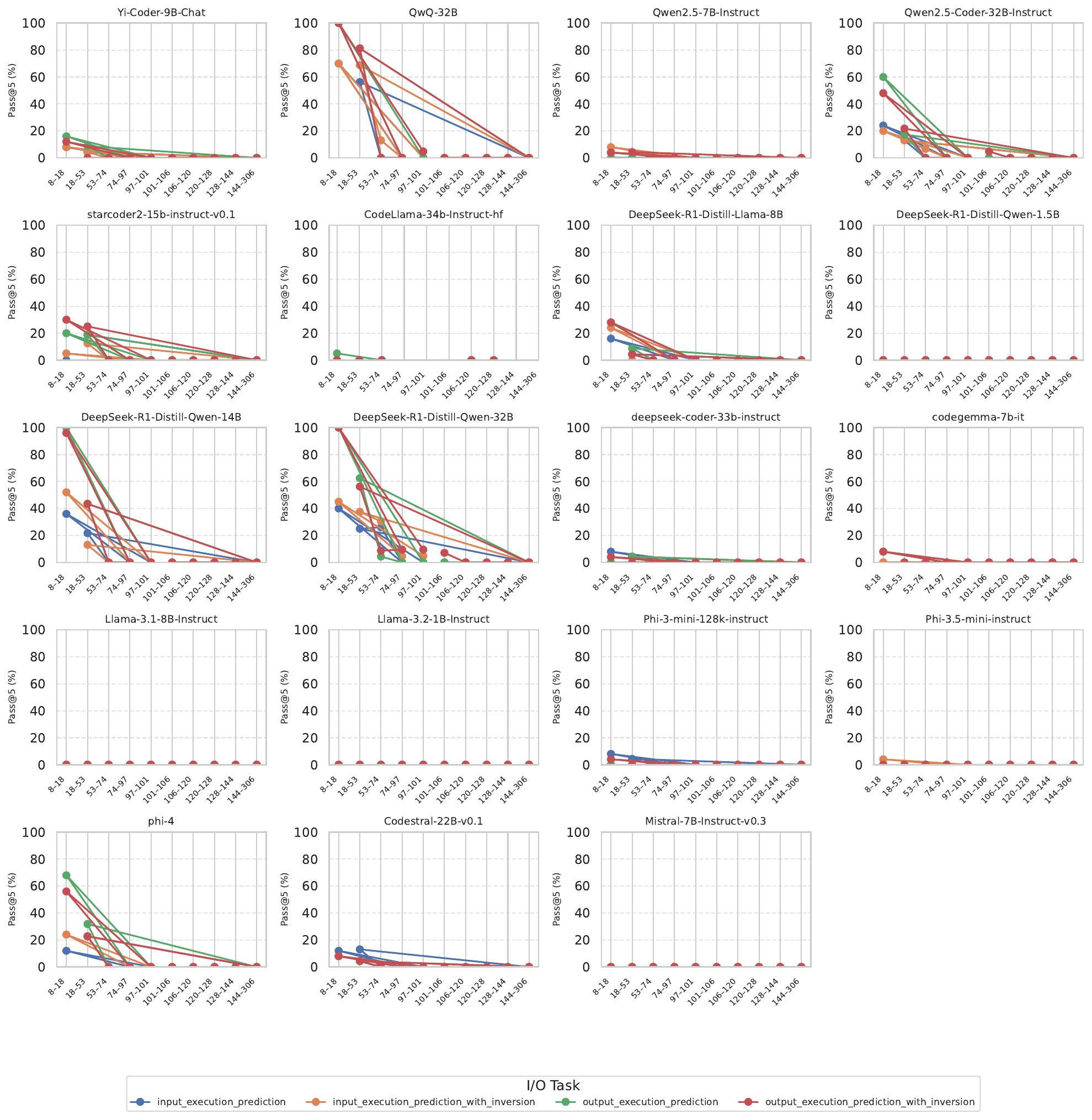}
    \caption{Input length vs.\ Pass@5 for LZW: the length-performance degradation is steeper than AE, consistent with LZW's growing dictionary making longer inputs exponentially harder to trace.}
    \label{fig:input-len-pass-5-lzw}
\end{figure*}

\begin{figure*}
    \centering
    \includegraphics[width=1\linewidth]{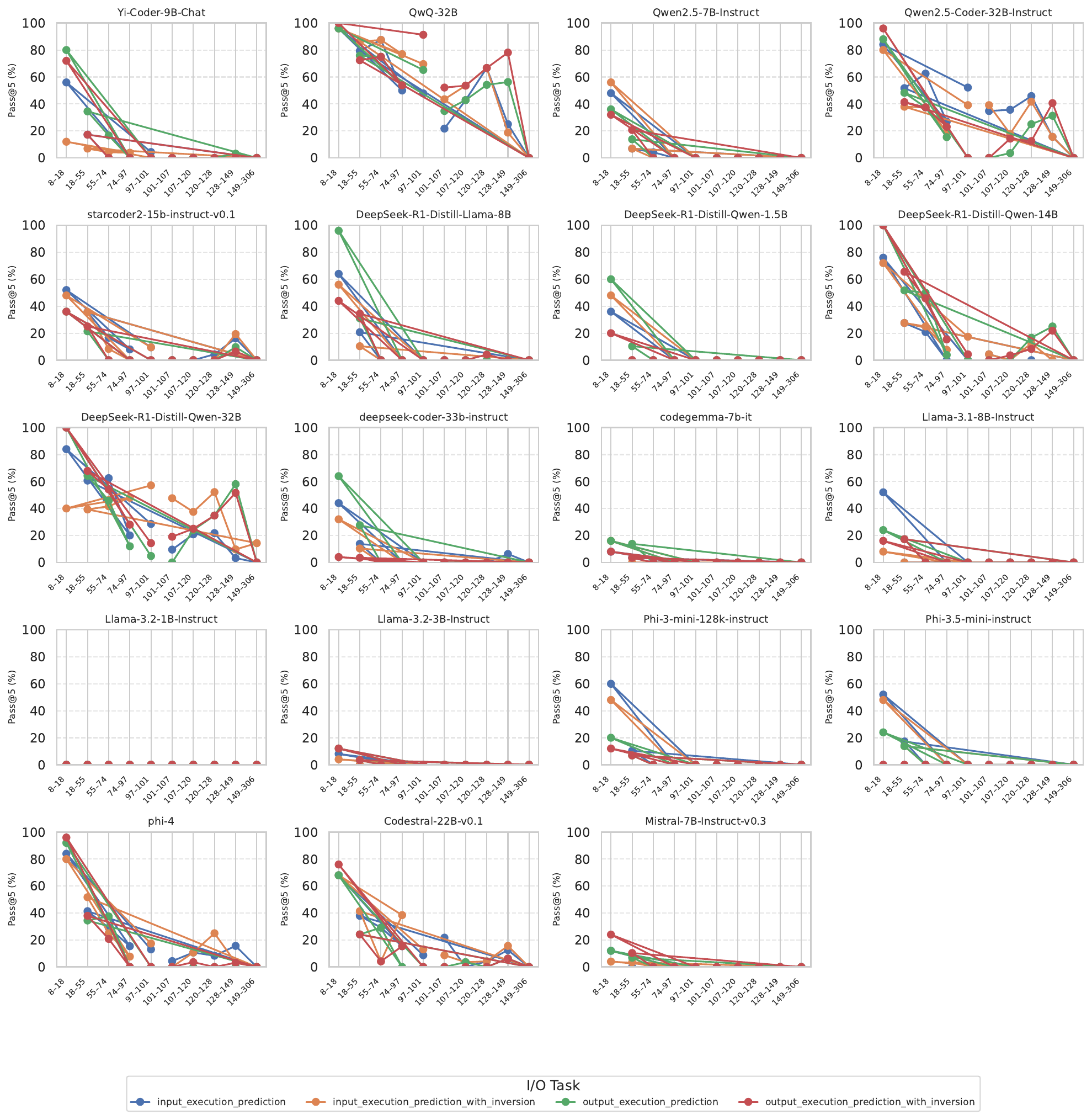}
    \caption{Input length vs.\ Pass@5 for RLE: performance is the most robust to length among all algorithms, yet still degrades for long inputs, particularly those dominated by non-repetitive character patterns.}
    \label{fig:input-len-pass-5-rle}
\end{figure*}

\begin{figure*}
    \centering
    \includegraphics[width=1\linewidth]{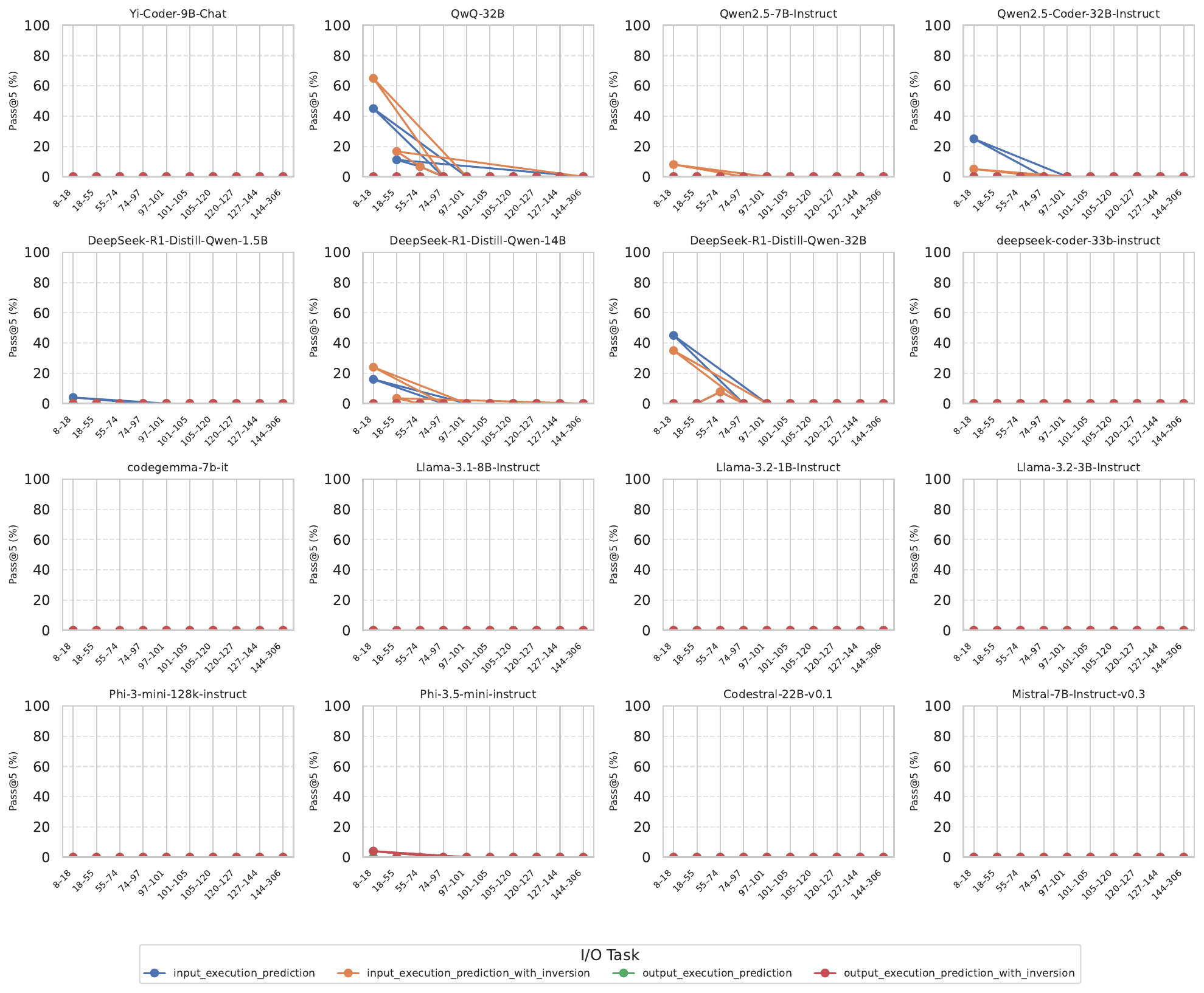}
    \caption{Input length vs.\ Pass@5 for Huffman: Pass@5 remains near zero across all length buckets, confirming that Huffman difficulty is driven by algorithmic complexity rather than input length alone.}
    \label{fig:input-len-pass-5-huffman}
\end{figure*}

Figures~\ref{fig:input-len-pass-5-ae}--\ref{fig:input-len-pass-5-huffman} show the relationship between input length and Pass@5 for all four compression algorithms. Each subplot corresponds to a model, with Pass@5 on the y-axis and input-length buckets on the x-axis. Across algorithms, longer inputs consistently yield lower Pass@5, confirming that input complexity is a key difficulty driver.

\section{Easy and difficult inputs for LLMs}
\label{sec:input-difficulty-level}
We present examples of the top-5 inputs that appear to be easy or difficult for LLMs across different compression algorithms. Specifically, \cref{tab:ae_top5_easiest_centered,tab:ae_top5_hardest_centered} report results for AE; \cref{tab:huffman_top5_easiest_centered,tab:huffman_top5_hardest_centered} for Huffman; \cref{tab:lzw_top5_easiest_centered,tab:lzw_top5_hardest_centered} for LZW; and \cref{tab:rle_top5_easiest_centered,tab:rle_top5_hardest_centered} for RLE.

\lstdefinestyle{tablerow}{
  basicstyle=\ttfamily\footnotesize,
  frame=none,
  aboveskip=0pt,
  belowskip=0pt,
  xleftmargin=0pt,
  xrightmargin=0pt,
  breaklines=true,
  breakatwhitespace=false,
  breakindent=0pt,
  columns=fullflexible,
  linewidth=\linewidth,
  keepspaces=true,
}

\newcolumntype{I}{>{\raggedright\arraybackslash}m{0.50\textwidth}}
\newcolumntype{N}{>{\centering\arraybackslash}m{0.10\textwidth}}

\begin{table*}[htbp]
\centering\footnotesize
\setlength{\tabcolsep}{5pt}
\begin{tabular}{I N N N}
\toprule
\textbf{Input} &
\makecell{\textbf{models}\\\textbf{(total)}} &
\makecell{\textbf{models}\\\textbf{(correct)}} &
\textbf{correct\_rate} \\
\midrule
\begin{minipage}[c]{0.48\textwidth}\vspace{4pt}\begin{lstlisting}[style=tablerow]
QWERTYUIOP
\end{lstlisting}\vspace{4pt}\end{minipage} & 20 & 10 & 0.50 \\
\midrule
\begin{minipage}[c]{0.48\textwidth}\vspace{4pt}\begin{lstlisting}[style=tablerow]
UUUUUUUU
\end{lstlisting}\vspace{4pt}\end{minipage} & 20 & 9 & 0.45 \\
\midrule
\begin{minipage}[c]{0.48\textwidth}\vspace{4pt}\begin{lstlisting}[style=tablerow]
IIIIIIIIIIII
\end{lstlisting}\vspace{4pt}\end{minipage} & 20 & 7 & 0.35 \\
\midrule
\begin{minipage}[c]{0.48\textwidth}\vspace{4pt}\begin{lstlisting}[style=tablerow]
POIUYTREWQ
\end{lstlisting}\vspace{4pt}\end{minipage} & 20 & 5 & 0.25 \\
\midrule
\begin{minipage}[c]{0.48\textwidth}\vspace{4pt}\begin{lstlisting}[style=tablerow]
ABABABAB
\end{lstlisting}\vspace{4pt}\end{minipage} & 20 & 4 & 0.20 \\
\bottomrule
\end{tabular}
\caption{AE: top-5 easiest inputs.}
\label{tab:ae_top5_easiest_centered}
\end{table*}

\begin{table*}[htbp]
\centering\footnotesize
\setlength{\tabcolsep}{5pt}
\begin{tabular}{I N N N}
\toprule
\textbf{Input} &
\makecell{\textbf{models}\\\textbf{(total)}} &
\makecell{\textbf{models}\\\textbf{(correct)}} &
\textbf{correct\_rate} \\
\midrule
\begin{minipage}[c]{0.48\textwidth}\vspace{4pt}\begin{lstlisting}[style=tablerow]
# Default values for mychart
---
replicaCount: 2
image:
  repository: nginx
  tag: stable-0
\end{lstlisting}\vspace{4pt}\end{minipage} & 20 & 0 & 0.00 \\
\midrule
\begin{minipage}[c]{0.48\textwidth}\vspace{4pt}\begin{lstlisting}[style=tablerow]
# Default values for mychart
---
replicaCount: 2
image:
  repository: nginx
  tag: stable-1
\end{lstlisting}\vspace{4pt}\end{minipage} & 20 & 0 & 0.00 \\
\midrule
\begin{minipage}[c]{0.48\textwidth}\vspace{4pt}\begin{lstlisting}[style=tablerow]
# Default values for mychart
---
replicaCount: 2
image:
  repository: nginx
  tag: stable-2
\end{lstlisting}\vspace{4pt}\end{minipage} & 20 & 0 & 0.00 \\
\midrule
\begin{minipage}[c]{0.48\textwidth}\vspace{4pt}\begin{lstlisting}[style=tablerow]
# Default values for mychart
---
replicaCount: 4
image:
  repository: nginx
  tag: stable-0
\end{lstlisting}\vspace{4pt}\end{minipage} & 20 & 0 & 0.00 \\
\midrule
\begin{minipage}[c]{0.48\textwidth}\vspace{4pt}\begin{lstlisting}[style=tablerow]
- hosts: all
  tasks:
    - name: ensure git0 installed
      apt:
        name: git0
        state: present
\end{lstlisting}\vspace{4pt}\end{minipage} & 20 & 0 & 0.00 \\
\bottomrule
\end{tabular}
\caption{AE: top-5 hardest inputs.}
\label{tab:ae_top5_hardest_centered}
\end{table*}

\begin{table*}[htbp]
\centering\footnotesize
\setlength{\tabcolsep}{5pt}
\begin{tabular}{I N N N}
\toprule
\textbf{Input} &
\makecell{\textbf{models}\\\textbf{(total)}} &
\makecell{\textbf{models}\\\textbf{(correct)}} &
\textbf{correct\_rate} \\
\midrule
\begin{minipage}[c]{0.48\textwidth}\vspace{4pt}\begin{lstlisting}[style=tablerow]
UUUUUUUU
\end{lstlisting}\vspace{4pt}\end{minipage} & 16 & 7 & 0.44 \\
\midrule
\begin{minipage}[c]{0.48\textwidth}\vspace{4pt}\begin{lstlisting}[style=tablerow]
ABABABAB
\end{lstlisting}\vspace{4pt}\end{minipage} & 16 & 6 & 0.38 \\
\midrule
\begin{minipage}[c]{0.48\textwidth}\vspace{4pt}\begin{lstlisting}[style=tablerow]
QWERTYUIOP
\end{lstlisting}\vspace{4pt}\end{minipage} & 16 & 6 & 0.38 \\
\midrule
\begin{minipage}[c]{0.48\textwidth}\vspace{4pt}\begin{lstlisting}[style=tablerow]
ABCDABCDABCDABCD
\end{lstlisting}\vspace{4pt}\end{minipage} & 16 & 4 & 0.25 \\
\midrule
\begin{minipage}[c]{0.48\textwidth}\vspace{4pt}\begin{lstlisting}[style=tablerow]
ABCDEFABCDEFABCDEF
\end{lstlisting}\vspace{4pt}\end{minipage} & 16 & 3 & 0.19 \\
\bottomrule
\end{tabular}
\caption{Huffman: top-5 easiest inputs.}
\label{tab:huffman_top5_easiest_centered}
\end{table*}

\begin{table*}[htbp]
\centering\footnotesize
\setlength{\tabcolsep}{5pt}
\begin{tabular}{I N N N}
\toprule
\textbf{Input} &
\makecell{\textbf{models}\\\textbf{(total)}} &
\makecell{\textbf{models}\\\textbf{(correct)}} &
\textbf{correct\_rate} \\
\midrule
\begin{minipage}[c]{0.48\textwidth}\vspace{4pt}\begin{lstlisting}[style=tablerow]
# Default values for mychart
---
replicaCount: 2
image:
  repository: nginx
  tag: stable-0
\end{lstlisting}\vspace{4pt}\end{minipage} & 16 & 0 & 0.00 \\
\midrule
\begin{minipage}[c]{0.48\textwidth}\vspace{4pt}\begin{lstlisting}[style=tablerow]
# Default values for mychart
---
replicaCount: 2
image:
  repository: nginx
  tag: stable-1
\end{lstlisting}\vspace{4pt}\end{minipage} & 16 & 0 & 0.00 \\
\midrule
\begin{minipage}[c]{0.48\textwidth}\vspace{4pt}\begin{lstlisting}[style=tablerow]
# Default values for mychart
---
replicaCount: 4
image:
  repository: nginx
  tag: stable-0
\end{lstlisting}\vspace{4pt}\end{minipage} & 16 & 0 & 0.00 \\
\midrule
\begin{minipage}[c]{0.48\textwidth}\vspace{4pt}\begin{lstlisting}[style=tablerow]
- hosts: all
  tasks:
    - name: ensure git1 installed
      apt:
        name: git1
        state: present
\end{lstlisting}\vspace{4pt}\end{minipage} & 16 & 0 & 0.00 \\
\midrule
\begin{minipage}[c]{0.48\textwidth}\vspace{4pt}\begin{lstlisting}[style=tablerow]
- hosts: all
  tasks:
    - name: ensure git2 installed
      apt:
        name: git2
        state: present
\end{lstlisting}\vspace{4pt}\end{minipage} & 16 & 0 & 0.00 \\
\bottomrule
\end{tabular}
\caption{Huffman: top-5 hardest inputs.}
\label{tab:huffman_top5_hardest_centered}
\end{table*}

\begin{table*}[htbp]
\centering\footnotesize
\setlength{\tabcolsep}{5pt}
\begin{tabular}{I N N N}
\toprule
\textbf{Input} &
\makecell{\textbf{models}\\\textbf{(total)}} &
\makecell{\textbf{models}\\\textbf{(correct)}} &
\textbf{correct\_rate} \\
\midrule
\begin{minipage}[c]{0.48\textwidth}\vspace{4pt}\begin{lstlisting}[style=tablerow]
QWERTYUIOP
\end{lstlisting}\vspace{4pt}\end{minipage} & 20 & 16 & 0.80 \\
\midrule
\begin{minipage}[c]{0.48\textwidth}\vspace{4pt}\begin{lstlisting}[style=tablerow]
POIUYTREWQ
\end{lstlisting}\vspace{4pt}\end{minipage} & 20 & 14 & 0.70 \\
\midrule
\begin{minipage}[c]{0.48\textwidth}\vspace{4pt}\begin{lstlisting}[style=tablerow]
FOX BROWN LAZY QUICK JUMPS
\end{lstlisting}\vspace{4pt}\end{minipage} & 20 & 10 & 0.50 \\
\midrule
\begin{minipage}[c]{0.48\textwidth}\vspace{4pt}\begin{lstlisting}[style=tablerow]
ITBVUUVBTI
\end{lstlisting}\vspace{4pt}\end{minipage} & 20 & 10 & 0.50 \\
\midrule
\begin{minipage}[c]{0.48\textwidth}\vspace{4pt}\begin{lstlisting}[style=tablerow]
MHYRFFRYHM
\end{lstlisting}\vspace{4pt}\end{minipage} & 20 & 10 & 0.50 \\
\bottomrule
\end{tabular}
\caption{LZW: top-5 easiest inputs.}
\label{tab:lzw_top5_easiest_centered}
\end{table*}

\begin{table*}[htbp]
\centering\footnotesize
\setlength{\tabcolsep}{5pt}
\begin{tabular}{I N N N}
\toprule
\textbf{Input} &
\makecell{\textbf{models}\\\textbf{(total)}} &
\makecell{\textbf{models}\\\textbf{(correct)}} &
\textbf{correct\_rate} \\
\midrule
\begin{minipage}[c]{0.48\textwidth}\vspace{4pt}\begin{lstlisting}[style=tablerow]
L(?OD4XezFnsc!ufFt-1A.-f3BDm:{K:wqA`^@&+qVC%
\end{lstlisting}\vspace{4pt}\end{minipage} & 20 & 0 & 0.00 \\
\midrule
\begin{minipage}[c]{0.48\textwidth}\vspace{4pt}\begin{lstlisting}[style=tablerow]
PESSEPPESPESSEPPESPESSEPPES
\end{lstlisting}\vspace{4pt}\end{minipage} & 20 & 0 & 0.00 \\
\midrule
\begin{minipage}[c]{0.48\textwidth}\vspace{4pt}\begin{lstlisting}[style=tablerow]
[2025-01-01 00:00:00] CRITICAL - DELETE /status by user:92f5df7b in 2.123s from 192.168.52.6 (server)
\end{lstlisting}\vspace{4pt}\end{minipage} & 20 & 0 & 0.00 \\
\midrule
\begin{minipage}[c]{0.48\textwidth}\vspace{4pt}\begin{lstlisting}[style=tablerow]
[2025-01-01 00:00:00] CRITICAL - GET /api/v1/order by user:b31022f0 in 0.792s from 192.168.111.238 (cache)
\end{lstlisting}\vspace{4pt}\end{minipage} & 20 & 0 & 0.00 \\
\midrule
\begin{minipage}[c]{0.48\textwidth}\vspace{4pt}\begin{lstlisting}[style=tablerow]
[2025-01-01 00:00:00] DEBUG - GET /metrics by user:018267c4 in 1.229s from 192.168.194.244 (worker)
\end{lstlisting}\vspace{4pt}\end{minipage} & 20 & 0 & 0.00 \\
\bottomrule
\end{tabular}
\caption{LZW: top-5 hardest inputs.}
\label{tab:lzw_top5_hardest_centered}
\end{table*}

\begin{table*}[htbp]
\centering\footnotesize
\setlength{\tabcolsep}{5pt}
\begin{tabular}{I N N N}
\toprule
\textbf{Input} &
\makecell{\textbf{models}\\\textbf{(total)}} &
\makecell{\textbf{models}\\\textbf{(correct)}} &
\textbf{correct\_rate} \\
\midrule
\begin{minipage}[c]{0.48\textwidth}\vspace{4pt}\begin{lstlisting}[style=tablerow]
POIUYTREWQ
\end{lstlisting}\vspace{4pt}\end{minipage} & 19 & 18 & 0.95 \\
\midrule
\begin{minipage}[c]{0.48\textwidth}\vspace{4pt}\begin{lstlisting}[style=tablerow]
QUICK JUMPS THE OVER LAZY
\end{lstlisting}\vspace{4pt}\end{minipage} & 19 & 18 & 0.95 \\
\midrule
\begin{minipage}[c]{0.48\textwidth}\vspace{4pt}\begin{lstlisting}[style=tablerow]
QWERTYUIOP
\end{lstlisting}\vspace{4pt}\end{minipage} & 19 & 18 & 0.95 \\
\midrule
\begin{minipage}[c]{0.48\textwidth}\vspace{4pt}\begin{lstlisting}[style=tablerow]
UUUUUUUU
\end{lstlisting}\vspace{4pt}\end{minipage} & 19 & 18 & 0.95 \\
\midrule
\begin{minipage}[c]{0.48\textwidth}\vspace{4pt}\begin{lstlisting}[style=tablerow]
CCCGGGCCCGGG
\end{lstlisting}\vspace{4pt}\end{minipage} & 19 & 17 & 0.89 \\
\bottomrule
\end{tabular}
\caption{RLE: top-5 easiest inputs.}
\label{tab:rle_top5_easiest_centered}
\end{table*}

\begin{table*}[htbp]
\centering\footnotesize
\renewcommand{\arraystretch}{1.0}
\setlength{\tabcolsep}{5pt}
\begin{tabular}{I N N N}
\toprule
\textbf{Input} &
\makecell{\textbf{models}\\\textbf{(total)}} &
\makecell{\textbf{models}\\\textbf{(correct)}} &
\textbf{correct\_rate} \\
\midrule
\begin{minipage}[c]{0.48\textwidth}\vspace{4pt}\begin{lstlisting}[style=tablerow]
?88dY1V7C[?)Xf,&qocnnse!+>5T^];S'5P QAZDVy}gt{^38E;'j~e'H'&j]`c4'a*7(l(v>S/h?jl%
\end{lstlisting}\vspace{4pt}\end{minipage} & 19 & 0 & 0.00 \\
\midrule
\parbox[c]{0.48\textwidth}{\vspace{4pt}\ttfamily\small\setlength{\baselineskip}{16pt}\seqsplit{AAACTCCGAATACCGATCCAGAGAGAGAGAGAGAGAGAGAGAGAGATCGACTGTCGCCGATGCGAT}\vspace{4pt}} & 19 & 0 & 0.00 \\
\midrule
\parbox[c]{0.48\textwidth}{\vspace{4pt}\ttfamily\small\setlength{\baselineskip}{16pt}\seqsplit{ATATACTTGGTCGGCGATCGAATAATAATAATAATAATAATAATAATAATAATAATAATAATAATAATAATCCGTCAATTAGGATCGCGTC}\vspace{4pt}} & 19 & 0 & 0.00 \\
\midrule
\parbox[c]{0.48\textwidth}{\vspace{4pt}\ttfamily\small\setlength{\baselineskip}{16pt}\seqsplit{ATGCCTGTCGTAGTAGTGGTCGTAGTCGTACGGTCAGCGGCAGCAGCCCCTCGGAAACCGGCTCCTTCACGGTCCCCTCGAGTGCCTGTTCCAGTGTCGTAG}\vspace{4pt}} & 19 & 0 & 0.00 \\
\midrule
\parbox[c]{0.48\textwidth}{\vspace{4pt}\ttfamily\small\setlength{\baselineskip}{16pt}\seqsplit{ATGCGGCCATCTCGGGCTGTTCCTGAGAAAGTCCCATCTGTCGCTCGGCCACCAGTACGGGTTGCTTCAGTAAAAGTTGCCGTGTCGTCATCGTCGCGAGTCCCAAAAGCCGCCCCGTCTCCGCGATCATCAGTGGTATAA}\vspace{4pt}} & 19 & 0 & 0.00 \\
\bottomrule
\end{tabular}
\caption{RLE: top-5 hardest inputs.}
\label{tab:rle_top5_hardest_centered}
\end{table*}

\section{Prompt difficulty for different data source categories}
\begin{figure*}[htbp]
  \centering
  \includegraphics[width=0.8\linewidth]{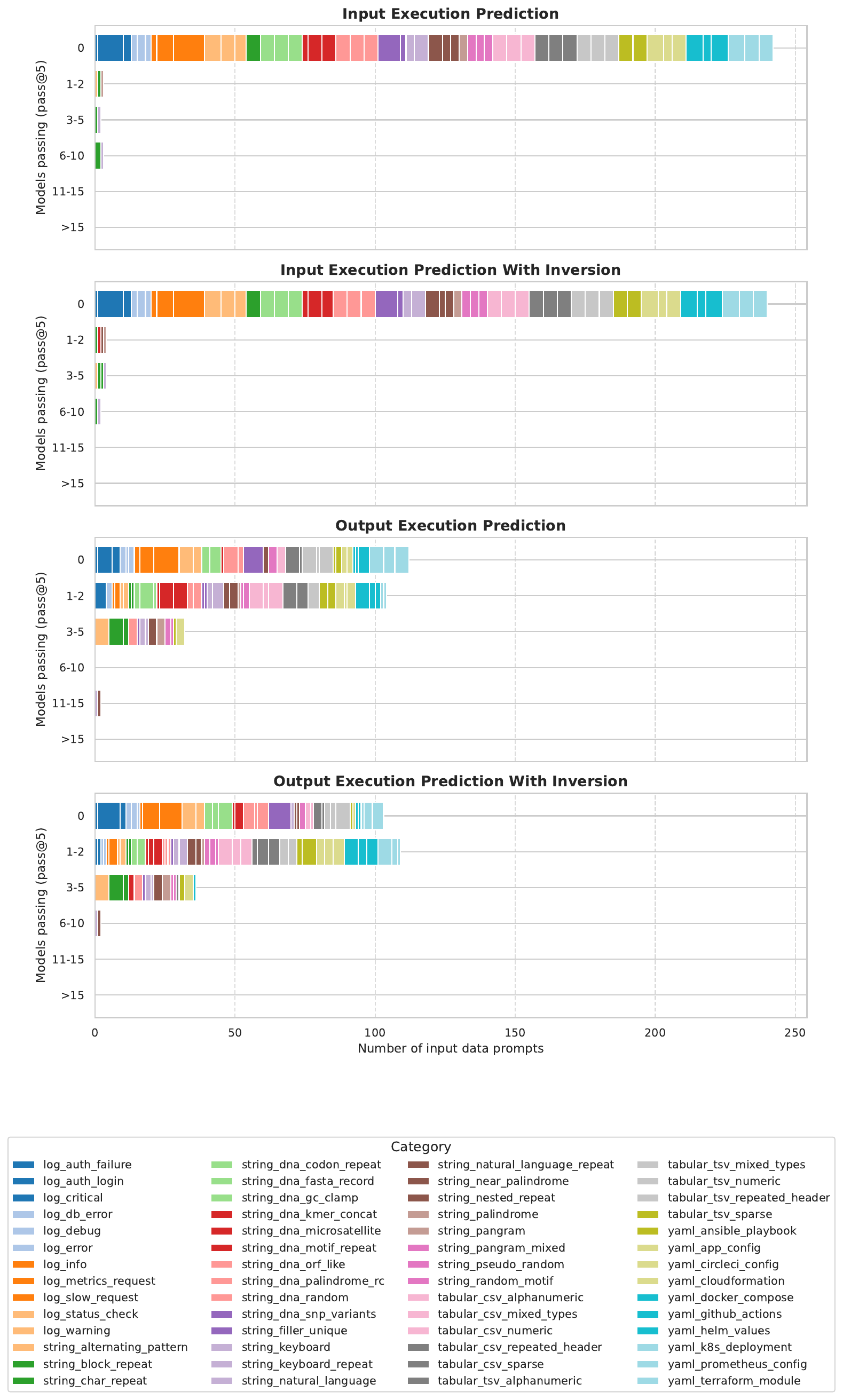}
  \caption{Prompt difficulty for AE: most prompts are solved by zero models, with short repetitive strings (e.g., keyboard sequences) forming the small solvable tail, while structured multi-line inputs such as YAML configs are universally failed.}
  \label{fig:prompt-difficulty-ae}
\end{figure*}

\begin{figure*}[htbp]
  \centering
  \includegraphics[width=0.8\linewidth]{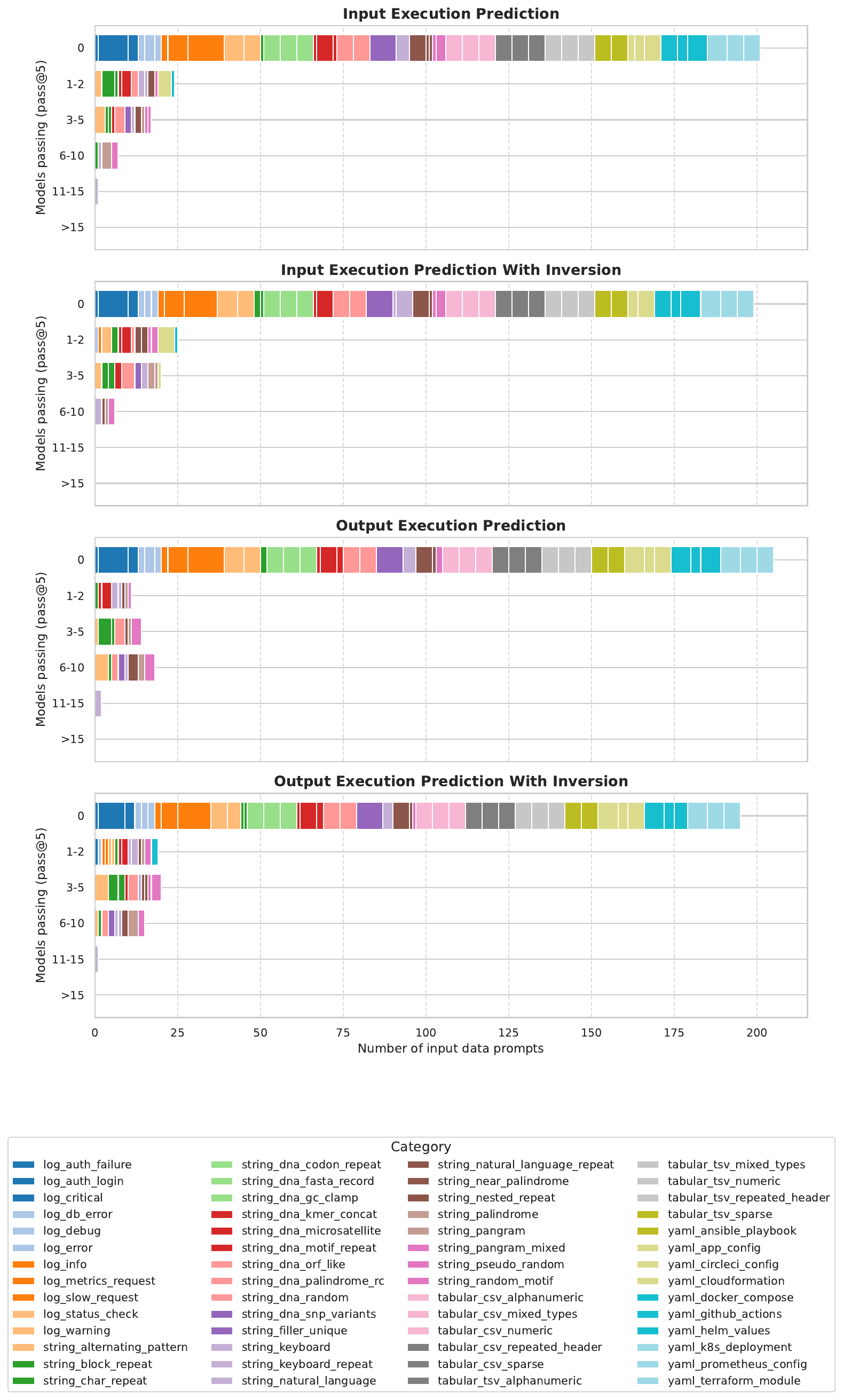}
  \caption{Prompt difficulty for LZW: dictionary-dependent encoding makes nearly all categories hard; only short, low-entropy strings like keyboard sequences are solved by a meaningful fraction of models, while log lines and config files remain universally unsolved.}
  \label{fig:prompt-difficulty-lzw}
\end{figure*}

\begin{figure*}[htbp]
  \centering
  \includegraphics[width=0.8\linewidth]{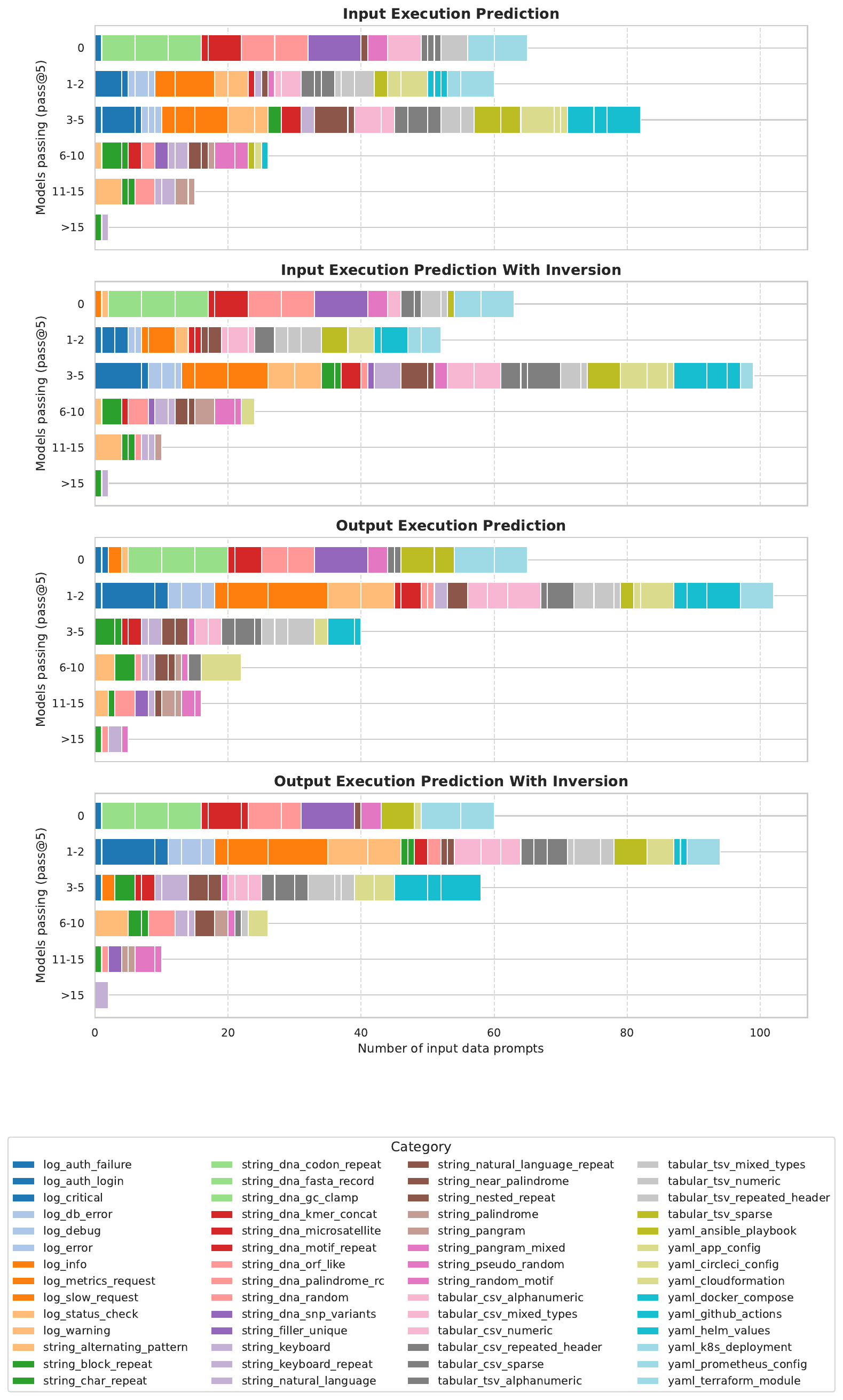}
  \caption{Prompt difficulty for RLE: the most solvable algorithm overall, with a larger right-heavy tail driven by simple repetitive strings; however, long DNA sequences and random character strings remain solved by very few models.}
  \label{fig:prompt-difficulty-rle}
\end{figure*}

\begin{figure*}[htbp]
  \centering
  \includegraphics[width=0.8\linewidth]{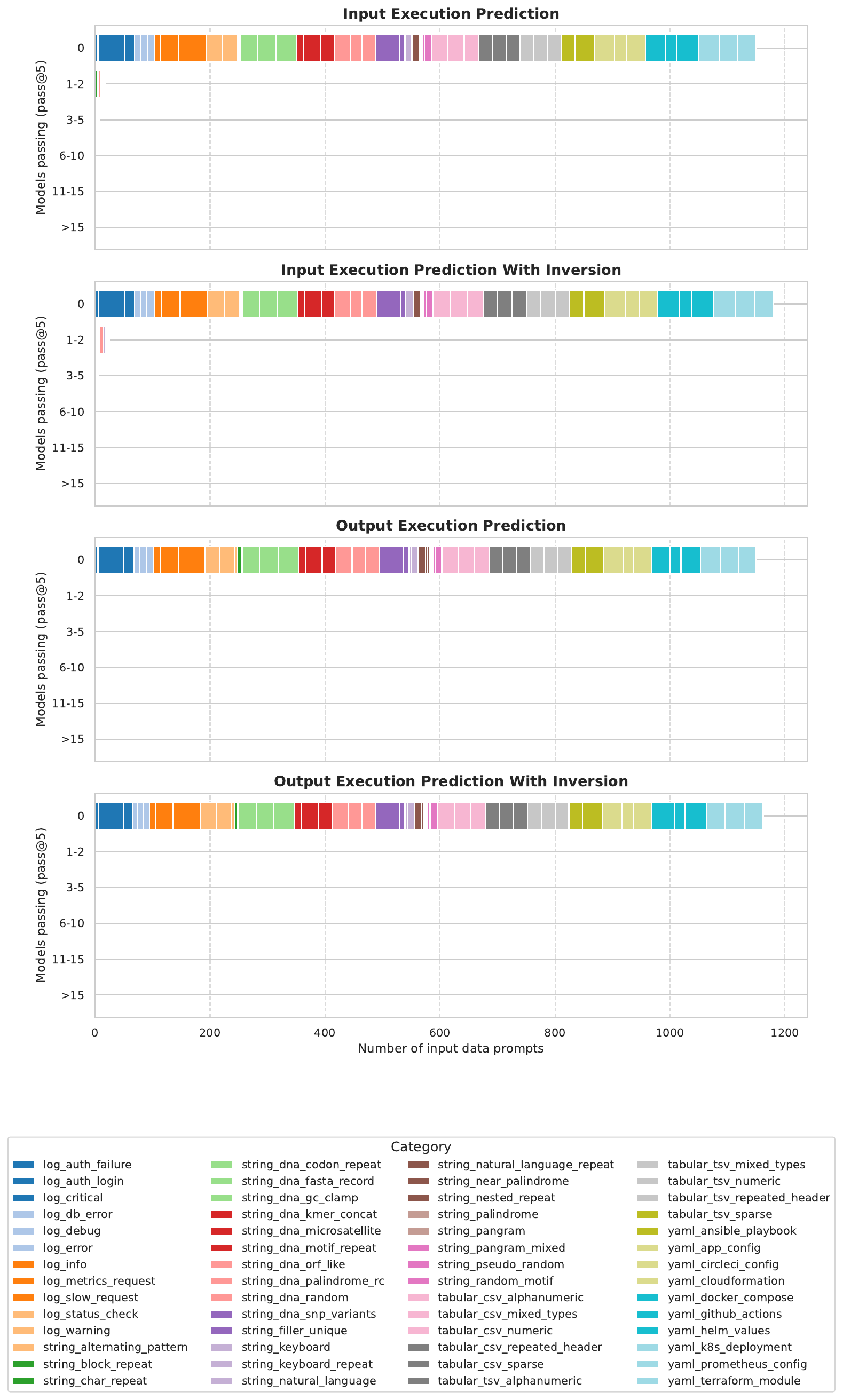}
  \caption{Prompt difficulty for Huffman: frequency-table construction makes this the hardest algorithm; nearly all prompts across all categories are solved by zero models, with only a small subset of short, high-repetition strings reaching partial success.}
  \label{fig:prompt-difficulty-huffman}
\end{figure*}

We use the Prompt Difficulty plot to show how challenging different data source categories are, with the x-axis indicating the number of models achieving Pass@5 and the y-axis showing the number of input strings. This breakdown by category highlights which prompt types are widely solved versus those that consistently challenge models. Figure~\ref{fig:prompt-difficulty-ae} shows for AE, Figure~\ref{fig:prompt-difficulty-lzw} shows for LZW, Figure~\ref{fig:prompt-difficulty-rle} shows for RLE, and Figure~\ref{fig:prompt-difficulty-huffman} shows for Huffman.

\section{Real-world significance}
Here are some real-world scenarios related to RTCE.

\begin{itemize}
    \item{\textbf{Code obfuscation and de-obfuscation:}} 
If the round-trip fails, the de-obfuscated code will contain errors, breaking the original program and demonstrating a failure to understand the underlying logic of the transformations.
    \item{\textbf{Data serialisation and deserialisation:}} If the LLM misunderstands even minor aspects of the schema (field names, integer vs. string encoding, escape sequences), the round-trip will not reconstruct the original data.
\end{itemize}

\end{document}